\documentclass{article} 
\usepackage{collas2023_conference,times}
\usepackage{easyReview}
\usepackage{booktabs}
\usepackage{subfigure}
\usepackage{colortbl}
\usepackage{tabularray}
\usepackage[export]{adjustbox}
\usepackage{stmaryrd}
\usepackage{float}

\usepackage{amsmath,amsfonts,bm}

\def\eqref#1{equation~\ref{#1}}

\def\1{\bm{1}}

\DeclareMathAlphabet{\mathsfit}{\encodingdefault}{\sfdefault}{m}{sl}
\SetMathAlphabet{\mathsfit}{bold}{\encodingdefault}{\sfdefault}{bx}{n}

\usepackage{hyperref}
\hypersetup{
    colorlinks=true,
    linkcolor=red,
    filecolor=magenta,
    urlcolor=blue,
    citecolor=purple,
    pdftitle={Overleaf Example},
    pdfpagemode=FullScreen,
    }

\newcommand{\model}[0]{ABIP}
\newcommand{\mpointnet}[0]{A-PointNet}
\newcommand{\saner}[0]{SANER}

\title{Evaluating Continual Learning on a Home Robot}

\author{Sam Powers \\
Robotics Institute \\
Carnegie Mellon University\\
\texttt{snpowers@cs.cmu.edu} \\
\And
Abhinav Gupta \\
Robotics Institute \\
Carnegie Mellon University\\
\texttt{abhinavg@cs.cmu.edu} \\
\And 
Chris Paxton  \\
Meta AI \\
\texttt{cpaxton@meta.com} \\
}

\newcommand{\github}{\url{https://github.com/AGI-Labs/continual_rl}}

\collasfinalcopy 

\begin{document}

\maketitle

\begin{abstract}
Robots in home environments need to be able to learn new skills continuously as data becomes available, becoming ever more capable over time while using as little real-world data as possible. However, traditional robot learning approaches typically assume large amounts of iid data, which is inconsistent with this goal.
In contrast, continual learning methods like CLEAR and SANE allow autonomous agents to learn off of a stream of non-iid samples; they, however, have not previously been demonstrated on real robotics platforms. In this work, we show how continual learning methods can be adapted for use on a real, low-cost home robot, and in particular look at the case where we have extremely small numbers of examples, in a task-id-free setting.
Specifically, we propose \saner{}, a method for continuously learning a library of skills, and \model{} (Attention-Based Interaction Policies) as the backbone to support it. We learn four sequential kitchen tasks on a low-cost home robot, using only a handful of demonstrations per task.

\end{abstract}

\section{Introduction}

For a home robot to be fully capable, it must be able to adapt to the changing needs of the home.
If a new appliance is purchased, the robot should be able to learn to use it seamlessly, without forgetting any previous skills it has learned. If new storage is purchased, the robot should be able to utilize previous knowledge it has about putting objects away to quickly learn to leverage it. Since no two homes, or humans, are the same, it is not feasible to rely entirely upon pre-training in controlled settings to enable a robot to do everything that might be asked of it.

In other words, we see the crucial need for in-home robots to utilize the methods of continual learning. In addition to the core continual learning goals highlighted in our examples -- mitigating {\textit{catastrophic forgetting}~\citep{mccloskey1989_catastrophic, ratcliff1990_connectionist}} and enabling {\textit{forward transfer}~\citep{lopezpaz2017_gem}} -- the robotics setting introduces several distinct challenges. 

Robotics is a challenging field for even a single task, as collecting data is time-intensive, {supervision is costly~\citep{pari2021_visualimitatation}}, and significant randomness can cause undesirable damage to both the home and the robot. Additionally, large amounts of resetting would be a {burden in a home setting~\citep{zhu2020_rlrobotics, eysenbach2017_leavenotrace}}. {Finally, the ability to generalize is critical since real-world settings are never exactly the same; sensor noise, error in robot control, shifts in lighting, and more all result in variation, even in otherwise static scenes.} As a result, learning on a real robot for a large enough set of tasks to validate new continuous learning methods has been out of reach. Existing work in the continual learning setting for robotics is limited and largely speculative~\citep{lesort2020continual}.

To resolve these issues, we propose \saner{}~\footnote{Code is available at \github}, an adaptation of the SANE algorithm~\citep{powers2022_sane} for the robotics imitation setting, which can be used to learn an ensemble of new skills given a handful of unstructured demonstrations of different tasks. {To be sufficiently general for robotics, \saner{} must use a policy that is sample-efficient and robust to noise. Additionally, \saner{} introduces several significant modifications to enable learning from imitation.}

{The policy we introduce with \saner{} is a} simple, highly sample-efficient point-cloud-based policy network we call Attention-Based Interaction Policies (\model{}). \model{} is based on PointNet++~\citep{qi2017_pointnet++}, and was designed to efficiently learn via imitation.
\model{} takes inspiration from prior work on perceptually-grounded action spaces for robots~\citep{zeng2018learning,zeng2021transporter,james2022coarse,shridhar2022perceiver,mo2021where2act,wu2021vat}, which has shown better data efficiency and robustness than directly learning to predict motor commands.
Using \model{}, our agents are capable of acquiring new skills with only two demonstrations, minimizing burden on the end-user and robot alike. {While we present \model{} as part of \saner{}, we believe that its sample efficiency and generalization capacity make it a useful component of any continual learning system for use with robotics.}



We evaluate \saner{} on four kitchen tasks using a low-cost mobile robot: two pick-and-place tasks and two tasks manipulating an articulated object. We demonstrate our method's ability to generalize to unseen object locations, unseen object types, and to clutter, all with very little data. We compare our new method to two other continual learning methods, modified to use \model{}: CLEAR~\citep{rolnick2018_clear} and EWC~\citep{kirkpatrick2016_ewc}. We demonstrate that \saner{} is more effective at both learning without forgetting and forward transfer in the robotics domain, and that \model{} is a useful component for learning general interaction policies from a small number of examples.

{
To summarize, in this paper we show for the first time how to modify existing state-of-the-art continual learning methods to a few-shot, real robot imitation learning domain, and describe the necessary algorithmic changes and evaluations. We also describe the requirements on a policy architecture which will make these experiments feasible, and provide a first version of such an architecture in \model{}.
}

{\begin{figure*}
\centering{
\includegraphics[width=0.8\textwidth]{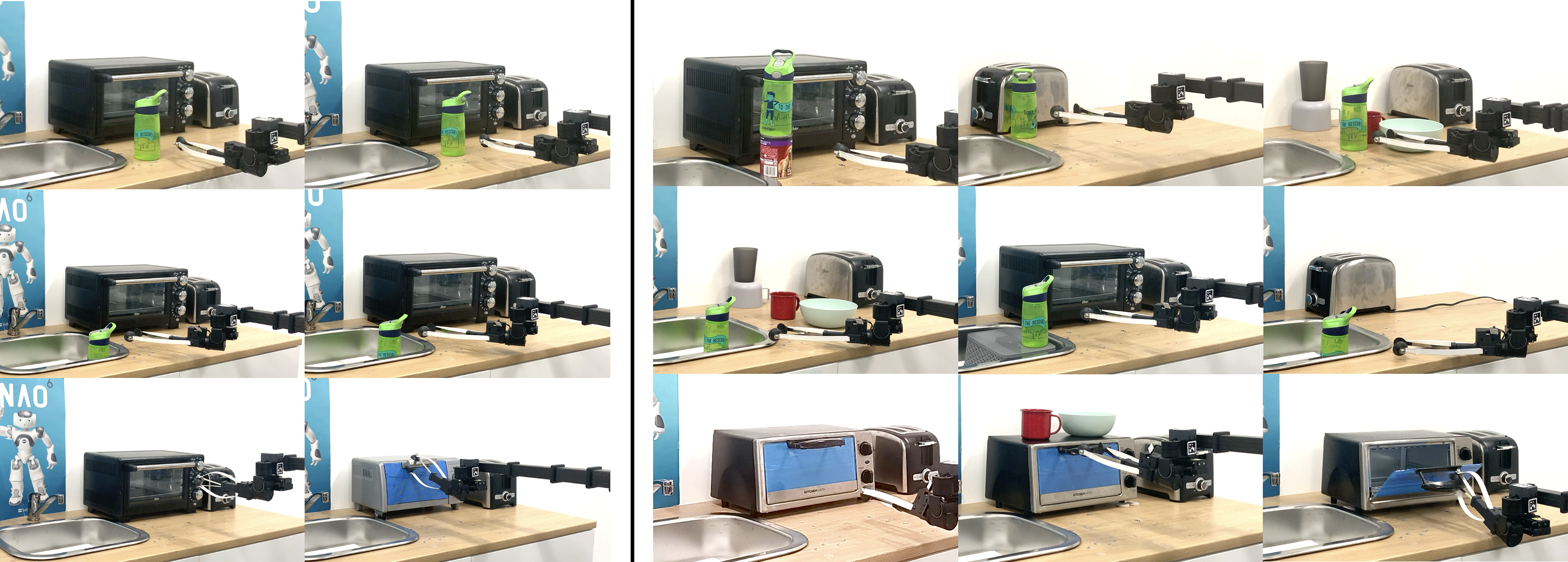} 
}
\caption{On the left we show the entire set of training settings for four tasks: picking and placing a bottle into and out of a sink, and opening and closing an oven. On the right are our evaluation settings. We show how continuous learning techniques like CLEAR~\citep{rolnick2018_clear} and SANE~\citep{powers2022_sane} can be extended to a low-data, learning-from-demonstration setting on a real robot, and can succeed on tasks that vary considerably from the original demonstrations.
}
\label{fig:cover}
\end{figure*}}


\vspace{-1em}
\section{Related Work}

Relevant prior work spans several fields, including continual learning~\citep{parisi2019_continualreview}, robotics, and imitation learning. We look both at prior work in continuous learning, reinforcement learning, and in learning generalizable robot policies.

\noindent \textbf{Continual learning for robotics.} Continual learning began as the study and mitigation of catastrophic forgetting~\citep{mccloskey1989_catastrophic, ratcliff1990_connectionist, robins1995_rehearsal}. 
It has since expanded to include forward transfer~\citep{lopezpaz2017_gem} and maintaining plasticity~\citep{mermillod2013_plasticity}, and more recently, generalization and sample efficiency~\citep{powers2022_cora}. In this work, we focus on generalization and sample efficiency {in the context of visuomotor policy learning}, in addition to forgetting, as these are critical pieces for robotics.

While a significant amount of work has focused on continual learning for supervised image classification~\citep{delange2019_classificationsurvey}, this too has expanded with time to include unsupervised learning~\citep{bagus2022_beyondsupervised} and reinforcement learning~\citep{kirkpatrick2016_ewc}. Continual learning in robotics has received some attention~\citep{lesort2020continual}, but while there have been a few works in simulation~\citep{wolczyk2021_continualworld, zentner2021_transferringskill, gaya2022_continualsubspace}, little has been put into practice. Key exceptions include work by ~\citet{rusu2017_progressivesim2real}, which uses Progressive Networks~\citep{rusu2016_progressive} to apply continual learning to the sim2real problem; \citet{gao2021_cril} utilize deep generative replay for the domain of continual imitation learning for robotics, using 15 demonstrations per real-world task, but their method requires using generative models to create whole new datasets in the new environment in order to avoid forgetting. {\citet{chens2022calable} propose an approach  which constructs policy mixtures for continuous control tasks, but can't learn visuomotor policies or generalize to different scenes.}

{One other route to continual learning for robotics would be, instead of learning visuomotor skills, to focus on grasp estimation and planning. In general, this sort of pipelined approach is brittle, failing due to occlusions or interference~\citep{kase2020transferable}, and existing open-set grasp estimation methods do not do task-oriented grasping~\citep{newbury2022deep}.}
{\citet{lomonaco2017core50} proposed a continuous object detection dataset, which could have some overlap for our methods, but uses only RGB data, which is far less useful for robust grasping~\citep{newbury2022deep}. Others have used continuous learning for object detection on real robots~\citep{ayub2020tell,ayub2022few}, but have not looked into reactive skill learning. However, ideas from continuous learning for object detection could be used in conjunction with SANER in the future to improve semantic learning and generalization.}

\noindent \textbf{Modular RL for robotics.} SANE, on which \saner{} is based, is an ensemble method based on the idea that having separate modules side-steps the problem of catastrophic forgetting. The idea of re-usable robotic skill libraries from demonstrations has been seen in ~\citet{tanneberg2021_skidraw, behbahani2021_episodicmemory, peng2019_composablepolicies, devin2016_modularmultitaskrobotics}, amongst many others~\citep{zentner2021_transferringskill}. Additionally, there has been work to apply modular methods to the continual robotics domain, such as in ~\citet{mendez2022_modular}, which looks at modular reinforcement learning in the simulated RoboSuite~\cite{zhu2020robosuite} environment, and~\citet{iwhiwhu2022_modulatingmasks}, which looks at lifelong reinforcement learning in a variety of simulated environments. These are relevant but not directly comparable to our work because much more data is available to train policies in these settings.
Others have noted zero-shot generalization as an important problem for RL, for example ~\cite{kirk2023survey} looks at avoiding overfitting to training environments through various means. In our case, we avoid overfitting through a mixture of augmentation and design of an attention-based, perception-driven robot policy.



\noindent \textbf{Perception-driven robot learning.} Recent work in robot learning focuses on building general, multi-purpose policies which can scale to a variety of scenarios based on sensor data. Much of this work has likewise focused on big-data settings. In Say-Can~\citep{ahn2022can} and RT-1~\citep{brohan2022rt}, for example, the authors use a large dataset to train a multi-task, language-conditioned transformer model. Given that these works require large amounts of data to train, they are not necessarily suitable for online teaching of robot skills in a home environment.

Conversely, there is a thread of work which functions on much smaller amounts of data but still achieves strong performance~\citep{zeng2018learning,zeng2021transporter,hundt2020good,james2022coarse,shridhar2022cliport,shridhar2022perceiver}. These works focus on a perceptual action space, where actions taken directly correlate with visual features, but often focus on a 2D vision of the world as a result~\citep{zeng2018learning,hundt2020good,zeng2021transporter,shridhar2022cliport}.

Recent work extended this sort of approach to deal with 3D environments more suitable for robotics tasks~\citep{shridhar2022perceiver}, but still uses very large transformer models that take a long time to train. An alternative is proposed by work like Where2Act~\citep{mo2021where2act} and VAT-Mart~\citep{wu2021vat}, which learn a point-cloud based policy which predicts an \textit{interaction point} and a corresponding trajectory. These are based on the powerful Pointnet++~\citep{qi2017_pointnet++} backbone, which provides convolutional-equivalent operators that work in 3d space. However, these policies are \textit{open-loop}, meaning that they cannot recover from failures. We build off this work, using a similar representation but modified to be slightly more general.

\section{Method}

\begin{figure}[bt]
    \centering
    \includegraphics[width=0.31\textwidth]{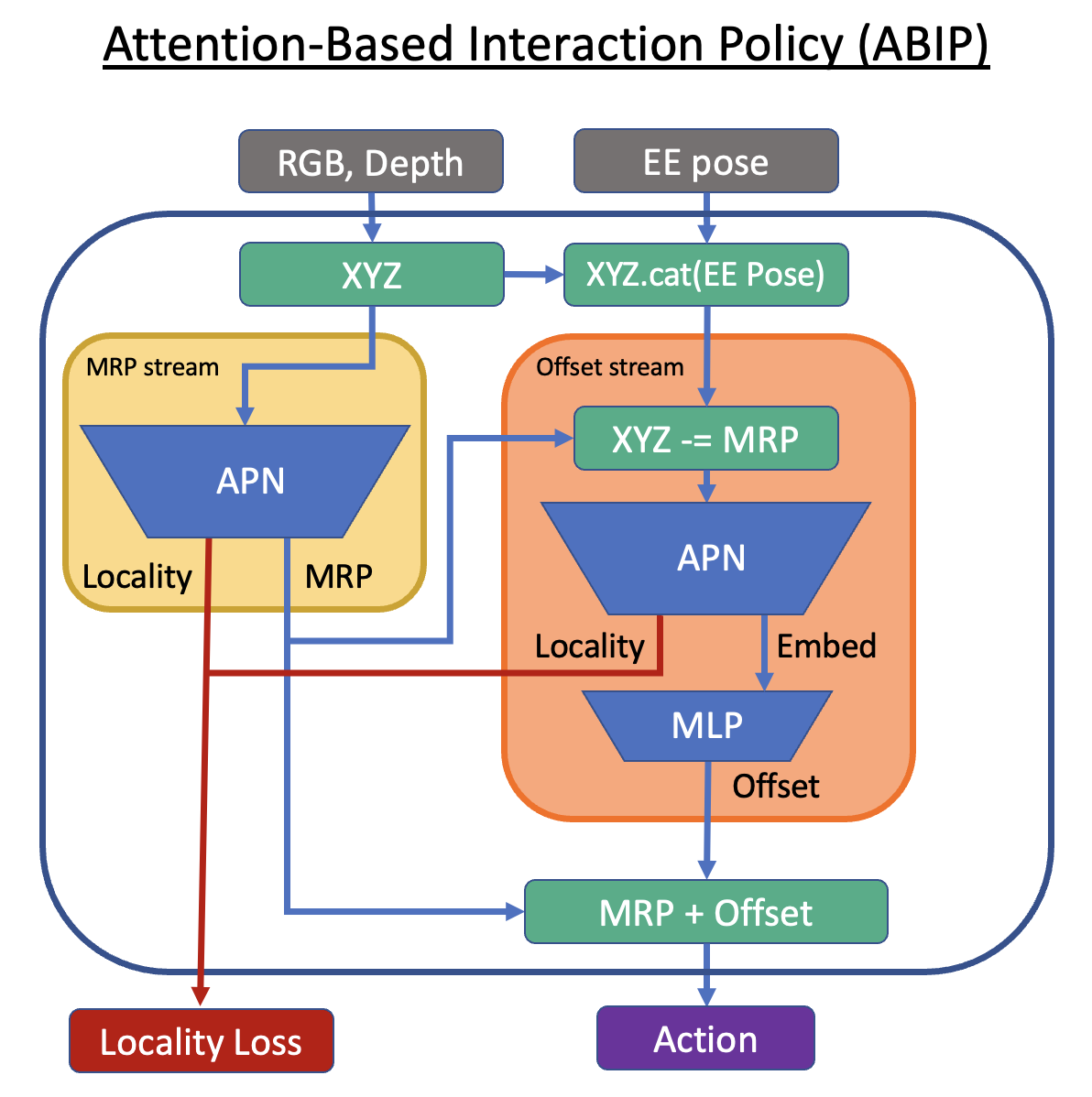}
    \raisebox{.8cm}{\includegraphics[width=0.35\textwidth]{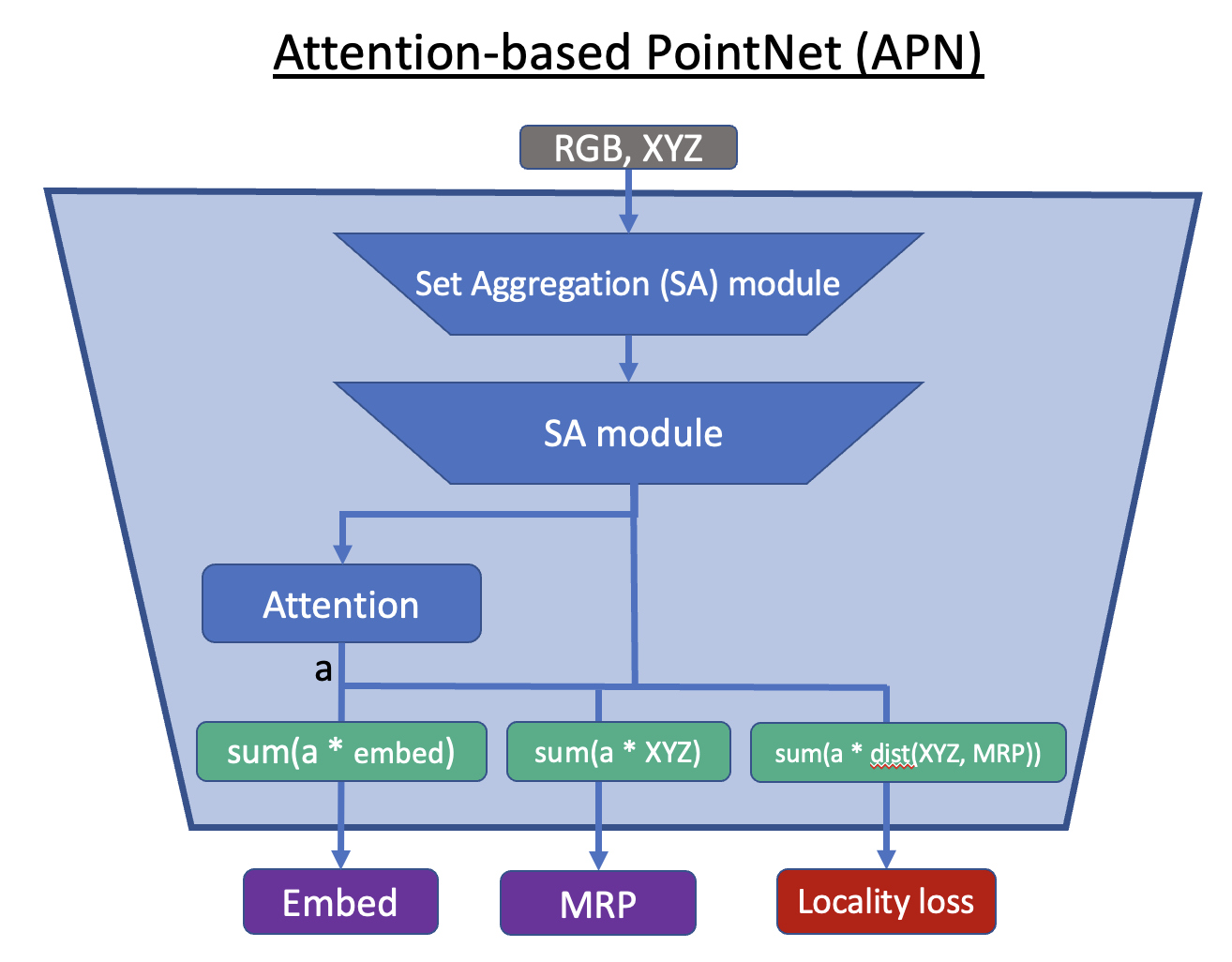}}
    \caption{
    Attention-based interaction policies (\model{}). We build an architecture which ignores variations and distractors by first predicting a \textit{Most Relevant Point} (MRP), and then predicting an offset from this point. \model{} is trained with a \textit{locality loss} which penalizes attending to points far from the robot's current position.
    }
    \label{fig:abip}
    \vskip -0.5cm
\end{figure}

We envision a home robot that learns an ever-growing library of skills. These skills could then be expanded, built upon, re-used, and perhaps even shared between agents in the future.
To achieve this, we propose \saner{}, an adaptation of SANE~\citep{powers2022_sane} for Robotics. 

SANE is an algorithm for the automatic creation and re-use of skill modules{; we provide an overview in Section \ref{sec:preliminaries}. To be suitable in the imitation learning setting, \saner{} makes several key changes. We break the necessary modifications up as follows: first, we discuss the new policy architecture used by each module, \textit{Attention-Based Interaction Policies} (\model{}), in Section \ref{sec:model}; second, we discuss changes to the critic in Section \ref{sec:saner_critic}; and third, we discuss changes to module creation in Section \ref{sec:saner_creation}.} 
Additionally, we modify two other continual learning methods, CLEAR and EWC, to use \model{} as well. We describe these adaptations in Section \ref{sec:existing_baselines}.

\vspace{-2em}
{\subsection{Preliminaries}}
\label{sec:preliminaries}
{\textbf{Problem Statement} Formally, we define an experiment as an ordered sequence of $N$ tasks, $(\mathcal{T}_0...\mathcal{T}_{N-1})$. During training, each task is specified by $k_{demo}$ demonstrations, where each demonstration is given as a trajectory of state, action pairs: $(\langle s, a \rangle)$. To allow us to assume the Markov property, we additionally augment $s$ with context $c$, taken as the state of the environment at the beginning of the episode. Each task is trained for $T$ timesteps, using points randomly sampled from the demonstration trajectories. Evaluation is performed by executing our learned policy on the real robot, in $k_{unseen}$ settings. More details on evaluation, including metric definitions, are provided in Section \ref{sec:evaluation}.}


{For the experiments presented in this paper,  the state is composed of an RGB-D image and the robot joint state, and the action is specified as the position and orientation (as a quaternion) of the end effector in world coordinates, plus the state of the gripper as a continuous variable in the range (-0.2, 0.2). Additionally, we include the fraction of the task completed in the action, which is used to determine the end of the episode. We use $N=4$, $k_{demo}=2$, $k_{unseen}$, and $T=10000$.}

\textbf{Overview of SANE} SANE is a task-id-free method that automatically detects and responds to drift in the setting by maintaining an \textit{ensemble} of models called \textit{nodes}. It does this by activating, merging, and creating modules automatically as follows: (1) Each module in its ensemble computes an activation score using a neural network we call the \textit{critic}; for a particular context, the node with the highest score is the one used. (2) As the node improves at the task, a stored version of the critic called the \textit{anchor} is updated. (3) A new node is created when the predicted activation score falls below the value predicted by the anchor by an amount that exceeds some allowed tolerance. This tolerance is called the \textit{uncertainty}, and is also predicted by the critic. 

{It is important to note that SANE uses CLEAR for each of its modules. CLEAR is a replay buffer method that uses reservoir sampling to give each sample an equal probability of remaining in the buffer, regardless of when it was collected. In the context of SANE, this helps the critic maintain a more accurate representation of the policy's capabilities. Since optimally a single task will activate a single SANE module, we use CLEAR for evaluate \model{} performance and single-task performance.}

\subsection{Attention-Based Interaction Policy (\model{})}
\label{sec:model}

Intuitively, in a cluttered and constantly changing setting like a human home, there will be many irrelevant details in the background of any demonstration that a user provides to \saner{}. Therefore, we split the action prediction problem into two steps: (1) we predict a \textit{Most Relevant Point}, or MRP, which tells us which region of the world the policy must attend to; and (2) we reactively predict \textit{actions} which determine where the robot should move in relation to that MRP: for example, how to approach the handle of an oven and when to close the gripper to grasp it. 


These two operations are performed sequentially using a modified PointNet++~\citep{qi2017_pointnet++} model that we refer to as \textit{Attention-based PointNet} (\mpointnet{}), shown in Figure~\ref{fig:abip}. The \textit{MRP Predictor} can then be agnostic to the position of the robot, instead focusing on the features of the object relevant to the overall task, while the \textit{Action Predictor} can learn to focus on features relevant just to what the next action should be. For example, in Figure~\ref{fig:attn_examples}, the MRP Predictor learns to focus on the handle; the Action Predictor focuses on the angle of the oven door. 

{\textbf{Image Pre-Processing}
First we convert the RGB and depth images into a point cloud.} We augment the point cloud of the current timestep with {our context $c$,} the point cloud from the beginning of the episode. This aids both in combating occlusion, as well as in disambiguating between similar observations that occur during different trajectories. {To reduce compute, we crop the working area to 1m, and down-sample using grid pooling, with a resolution of 1cm for the current timestep and 2.5cm for the context. Specifically, we select a random point in each voxel, to reduce overfitting.}

\subsubsection{Attention-Based PointNet (\mpointnet{})}

{Our Attention-Based PointNet module (\mpointnet{}), a core component of both our MRP and Action Prediction streams, augments PointNet++ via the addition of an attention mechanism, allowing us to learn to ignore irrelevant details of the scene.}

In more detail, \mpointnet{} passes the point cloud through two set aggregation modules that process information at different scales, as in PointNet++~\citep{qi2017_pointnet++}. This allows us to harness both structural information from the scene as well as the finer details necessary for manipulation. Further detail on these networks is provided in Appendix \ref{appendix:network_details}.



The output from the second {set aggregation} module~\citep{qi2017_pointnet++} is a reduced point cloud $P$. We concatenate the embedding, $m_i$, and position, $p_i$, of each point $i$ and pass these into an attention network, which is an MLP with output dimension 1, and softmax the resulting values to compute the attention value, $a_i$. This attention is then used to produce both a weighted average position, $\bar{p} = \sum_{i \in P} a_i p_i$ and a weighted average embedding,  $\bar{m} = \sum_{i \in P} a_i m_i$.

\begin{figure}[ht]
    \centering
    \subfigure[]{
        \includegraphics[width=0.15\textwidth,trim=40em 10em 40em 5em, clip]{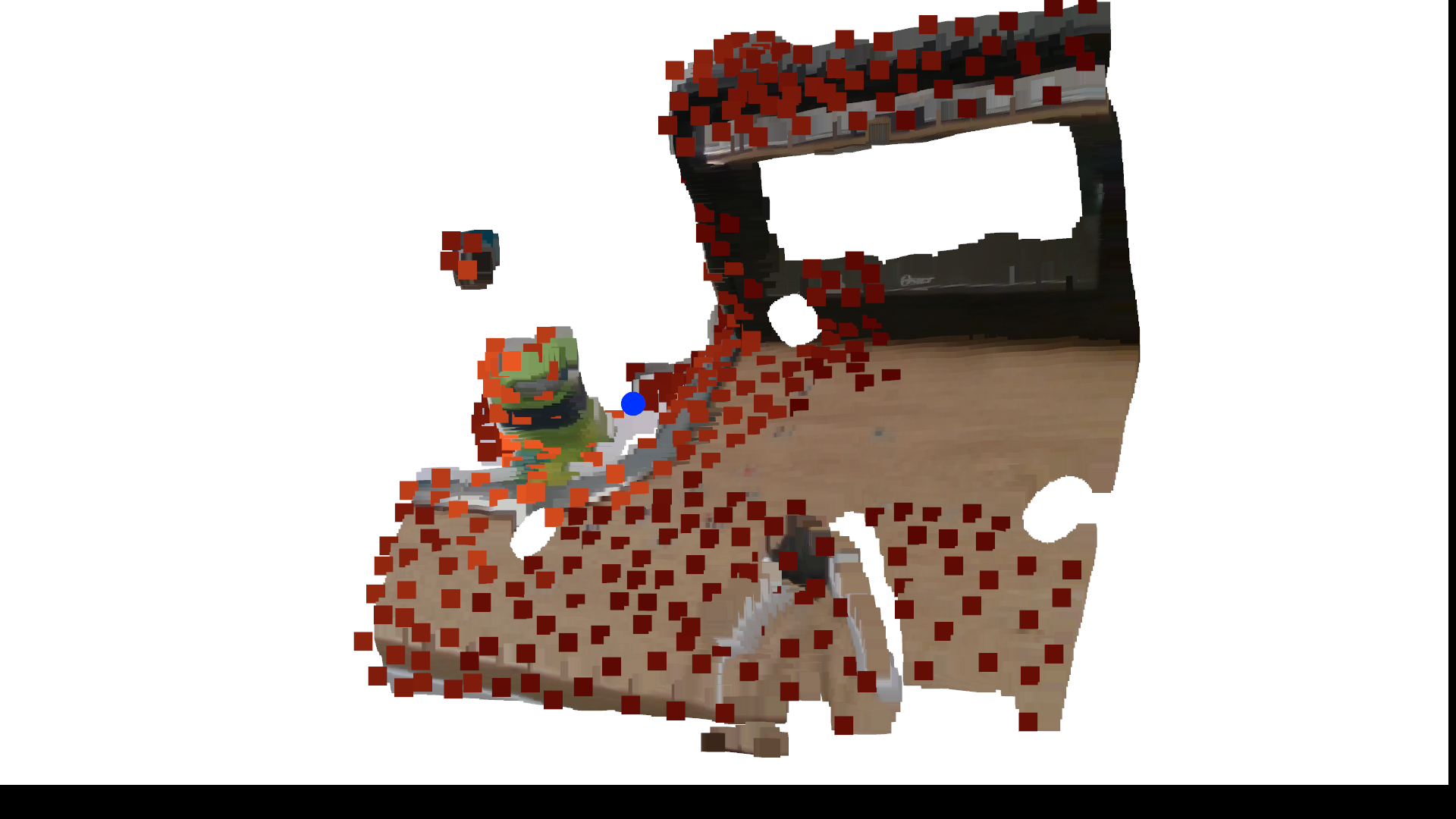}
        
    }
    \subfigure[]{
        \includegraphics[width=0.15\textwidth,trim=50em 10em 35em 5em, clip]{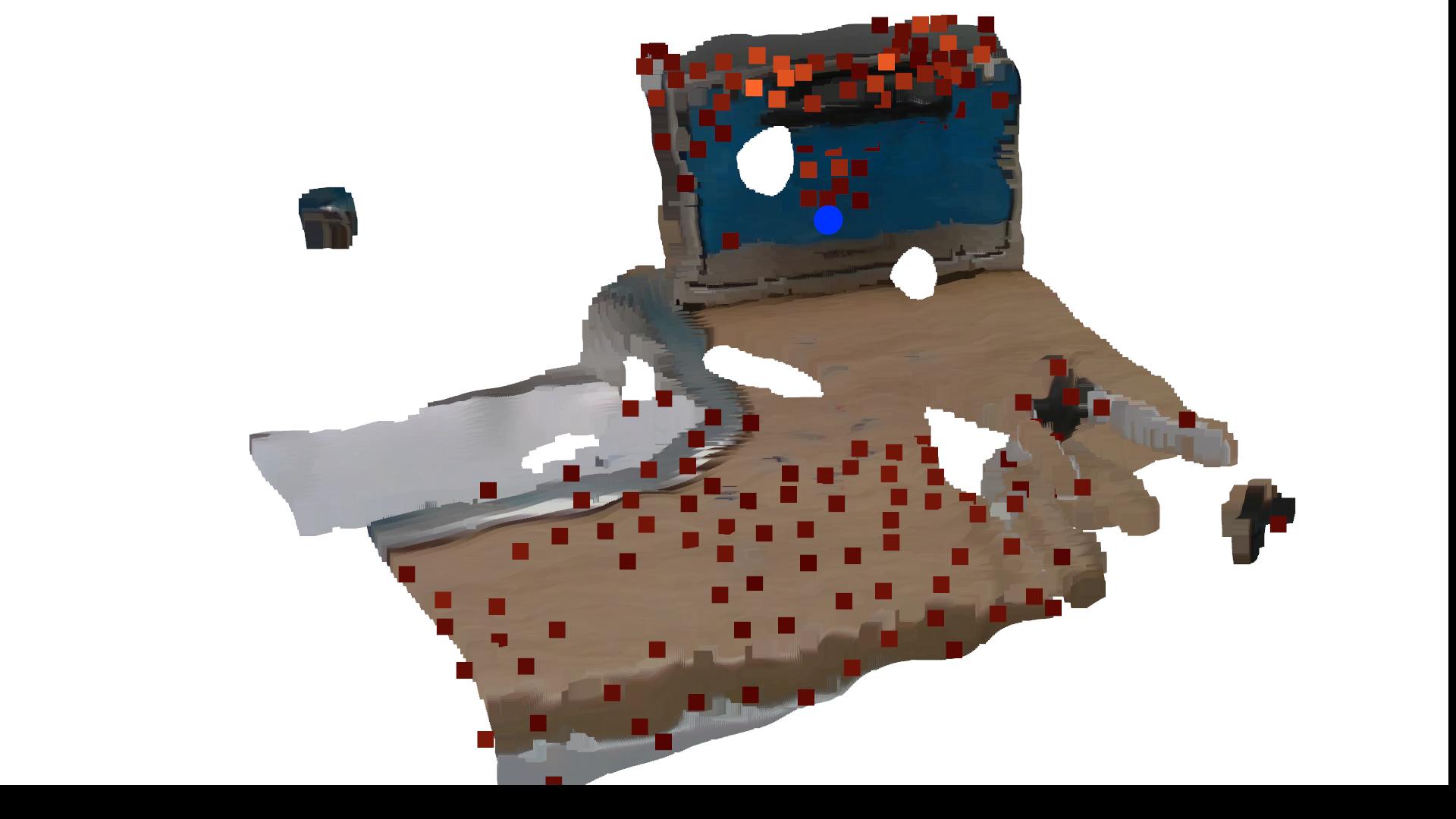}
    }
    \subfigure[]{
        \includegraphics[width=0.15\textwidth,trim=50em 10em 35em 5em, clip]{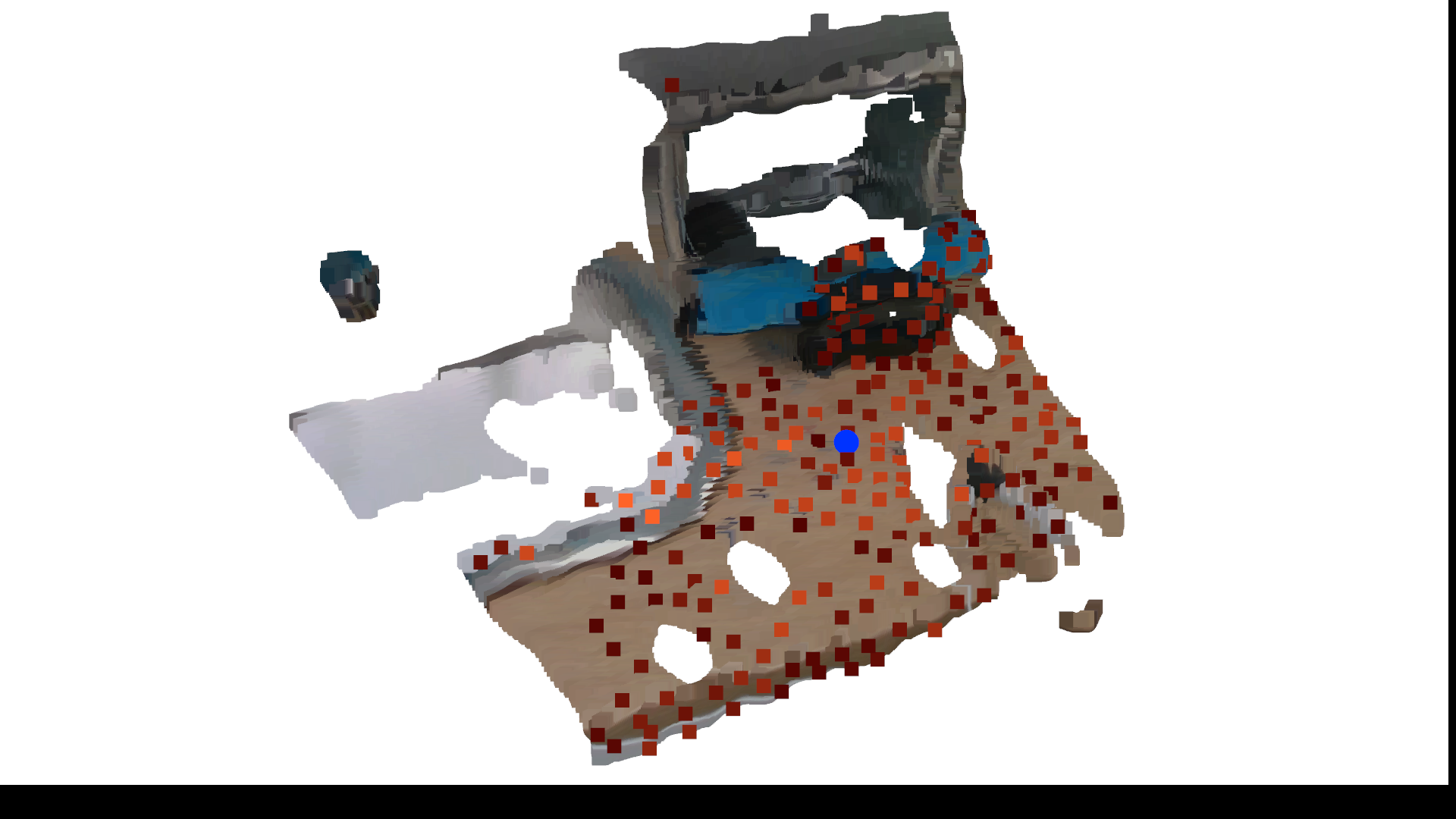}
    }

    \subfigure[]{
        \includegraphics[width=0.15\textwidth,trim=40em 10em 40em 5em, clip]{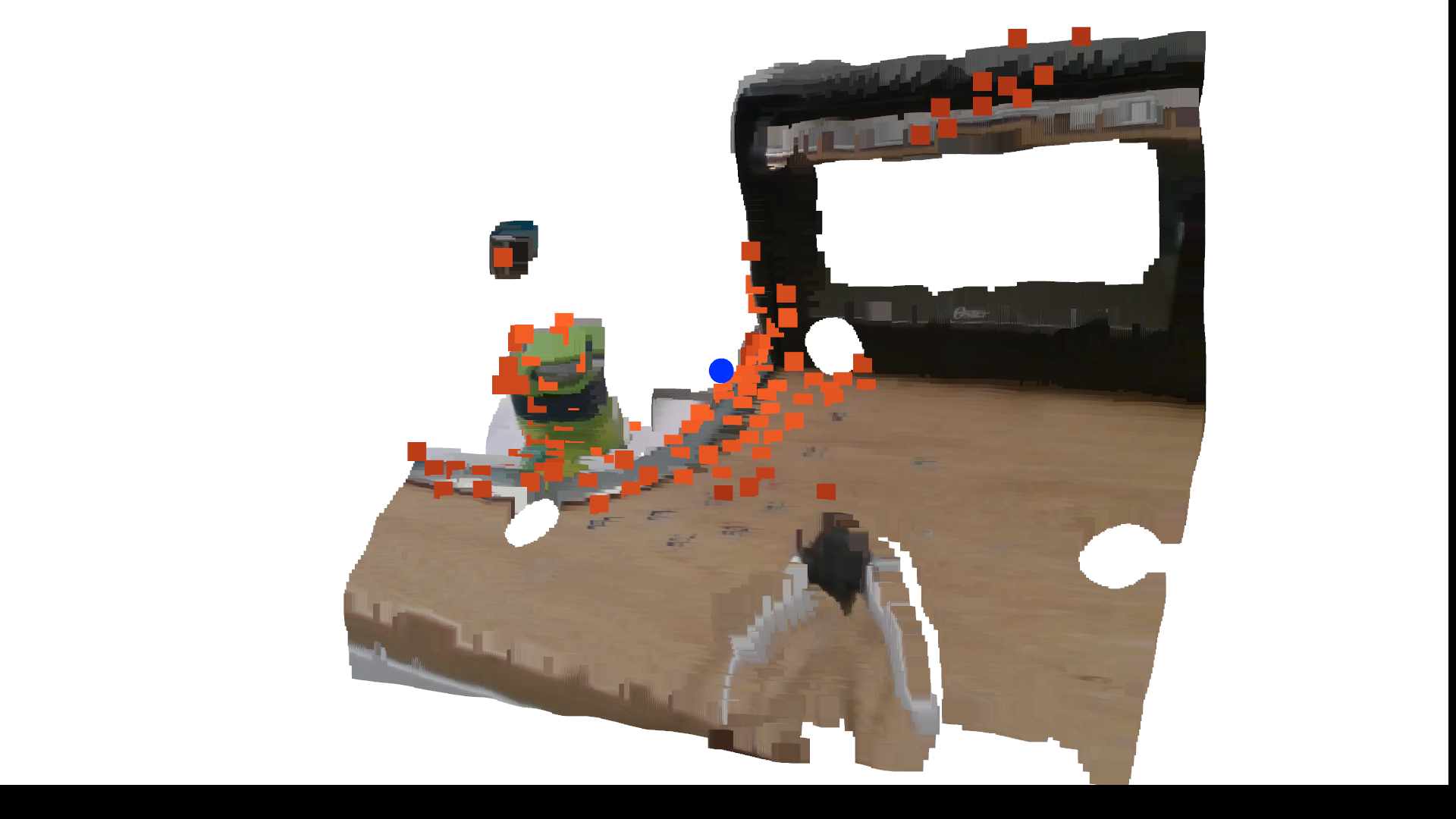}
    }
    \subfigure[]{
        \includegraphics[width=0.15\textwidth,trim=50em 10em 35em 5em, clip]{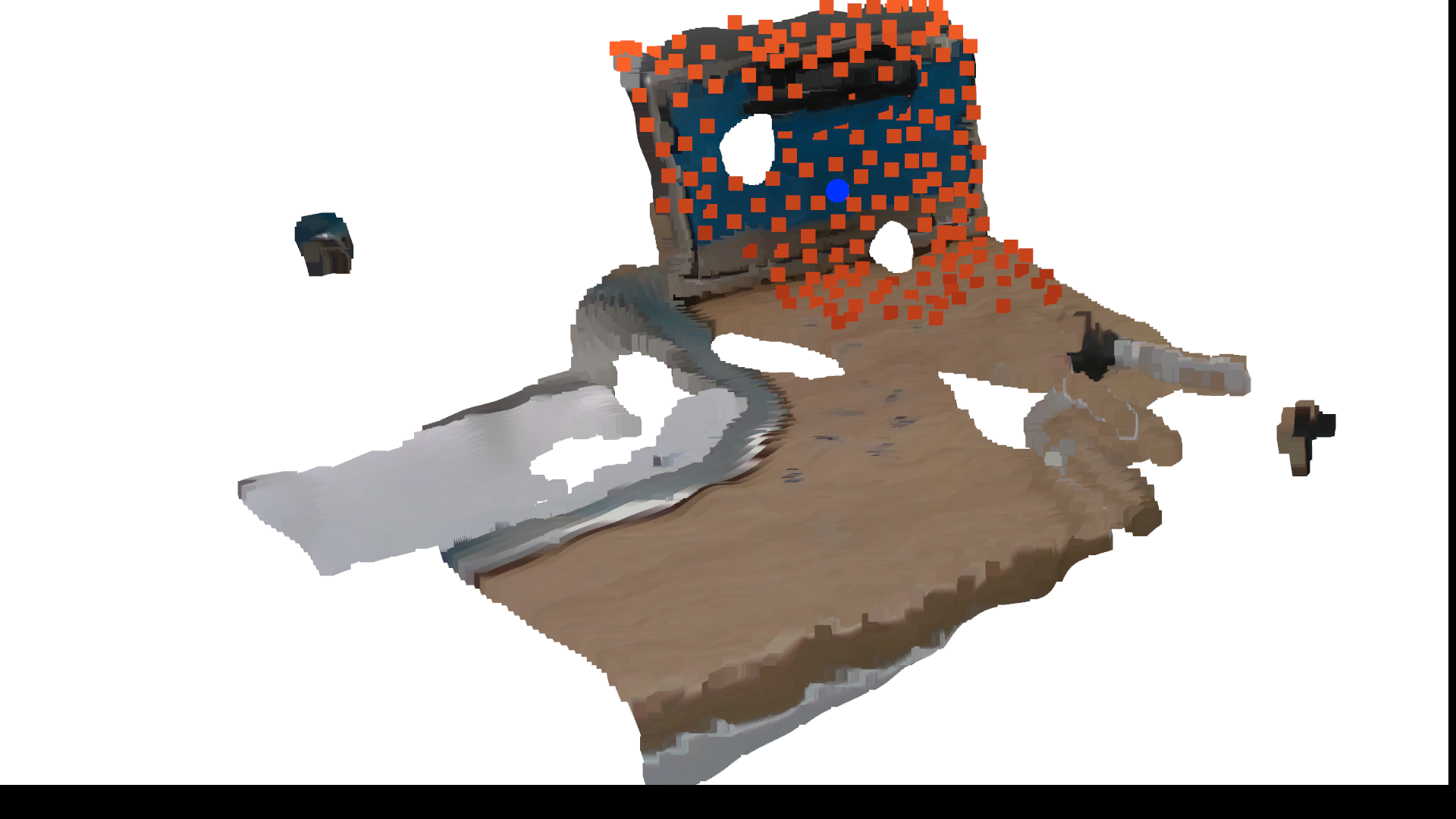}
    }
    \subfigure[]{
        \includegraphics[width=0.15\textwidth,trim=50em 10em 35em 5em, clip]{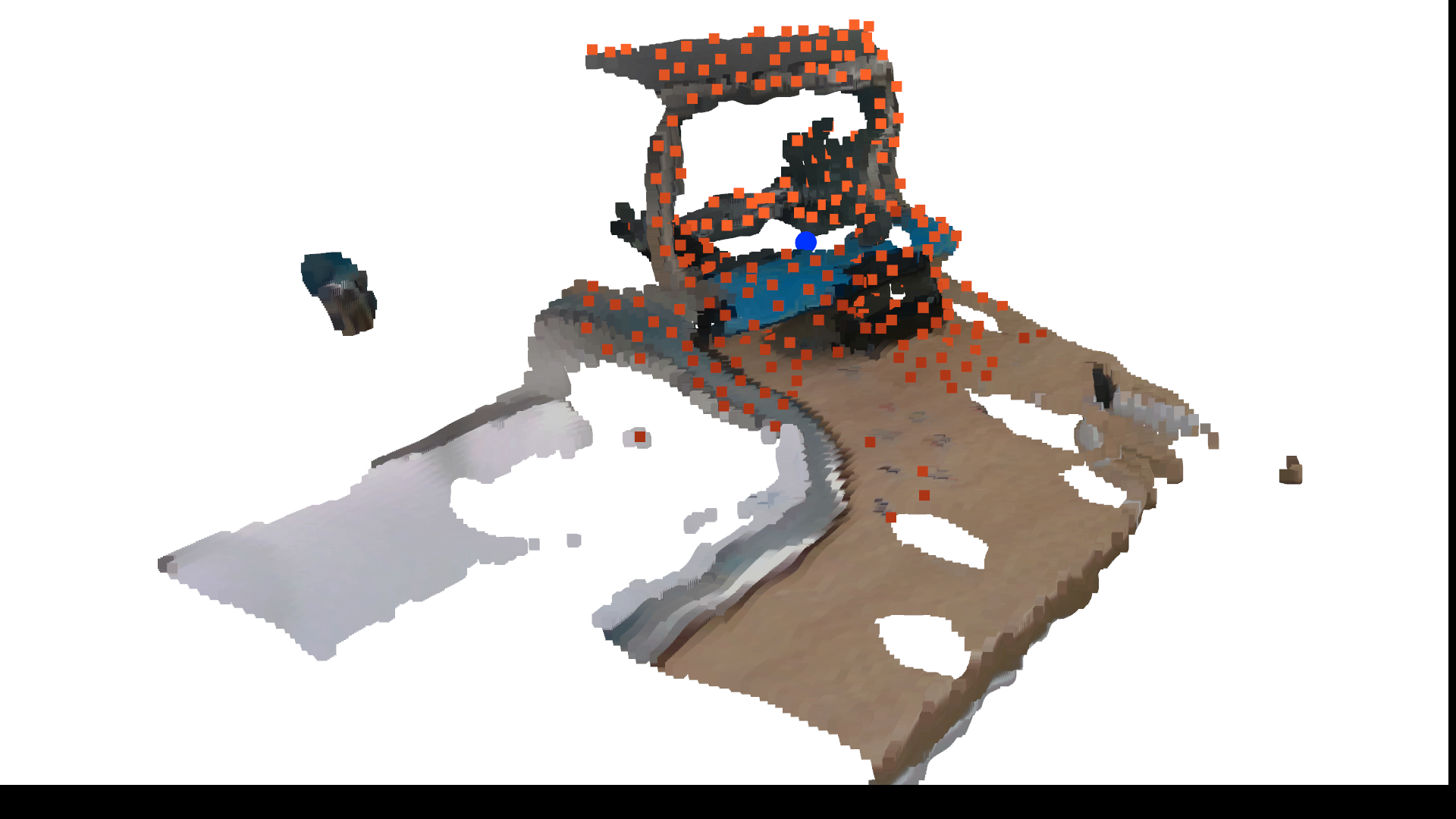}
    }
    
    \subfigure[]{
        \includegraphics[width=0.15\textwidth,trim=40em 10em 40em 5em, clip]{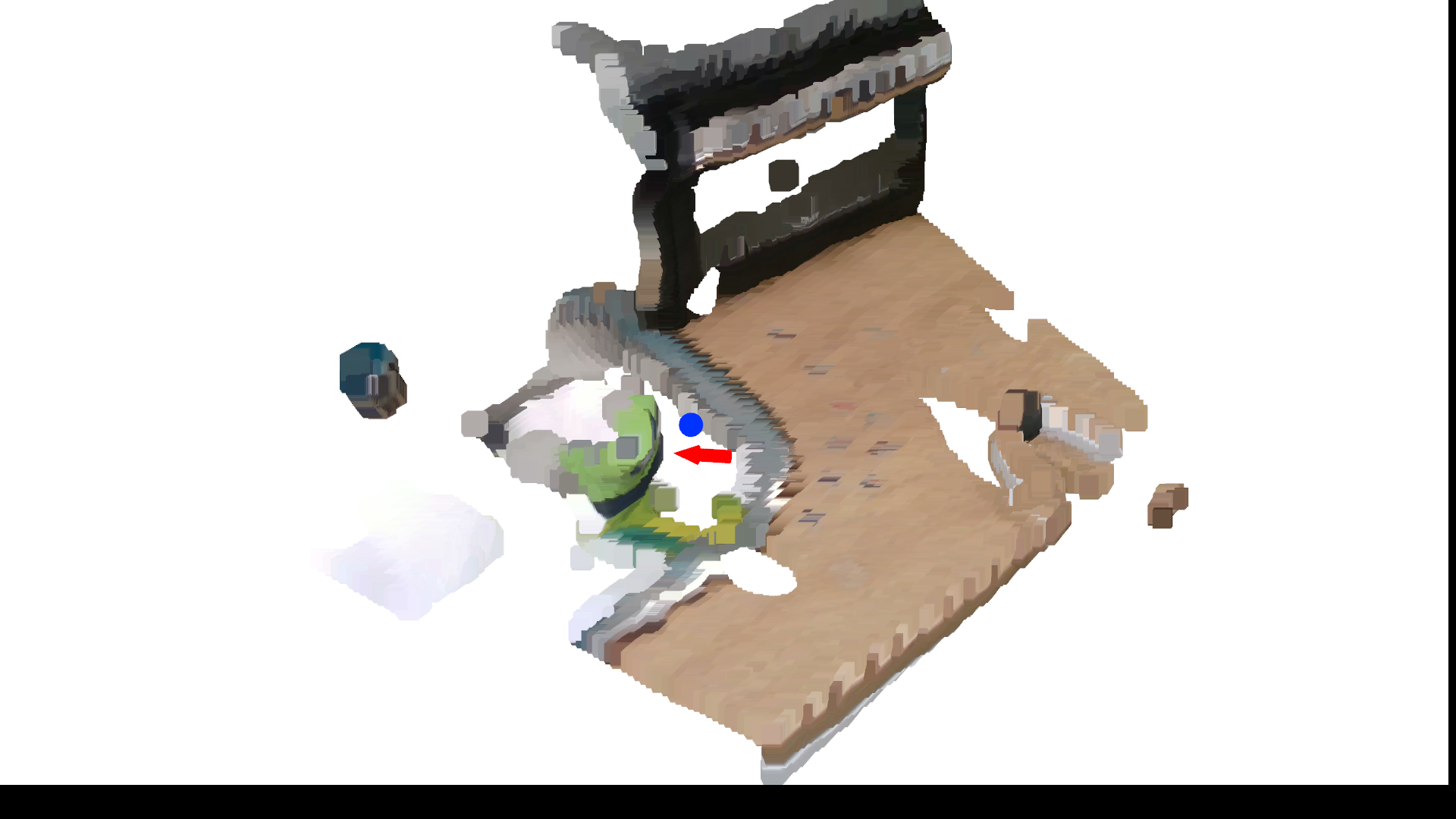}
    }
    \subfigure[]{
        \includegraphics[width=0.15\textwidth,trim=50em 10em 35em 5em, clip]{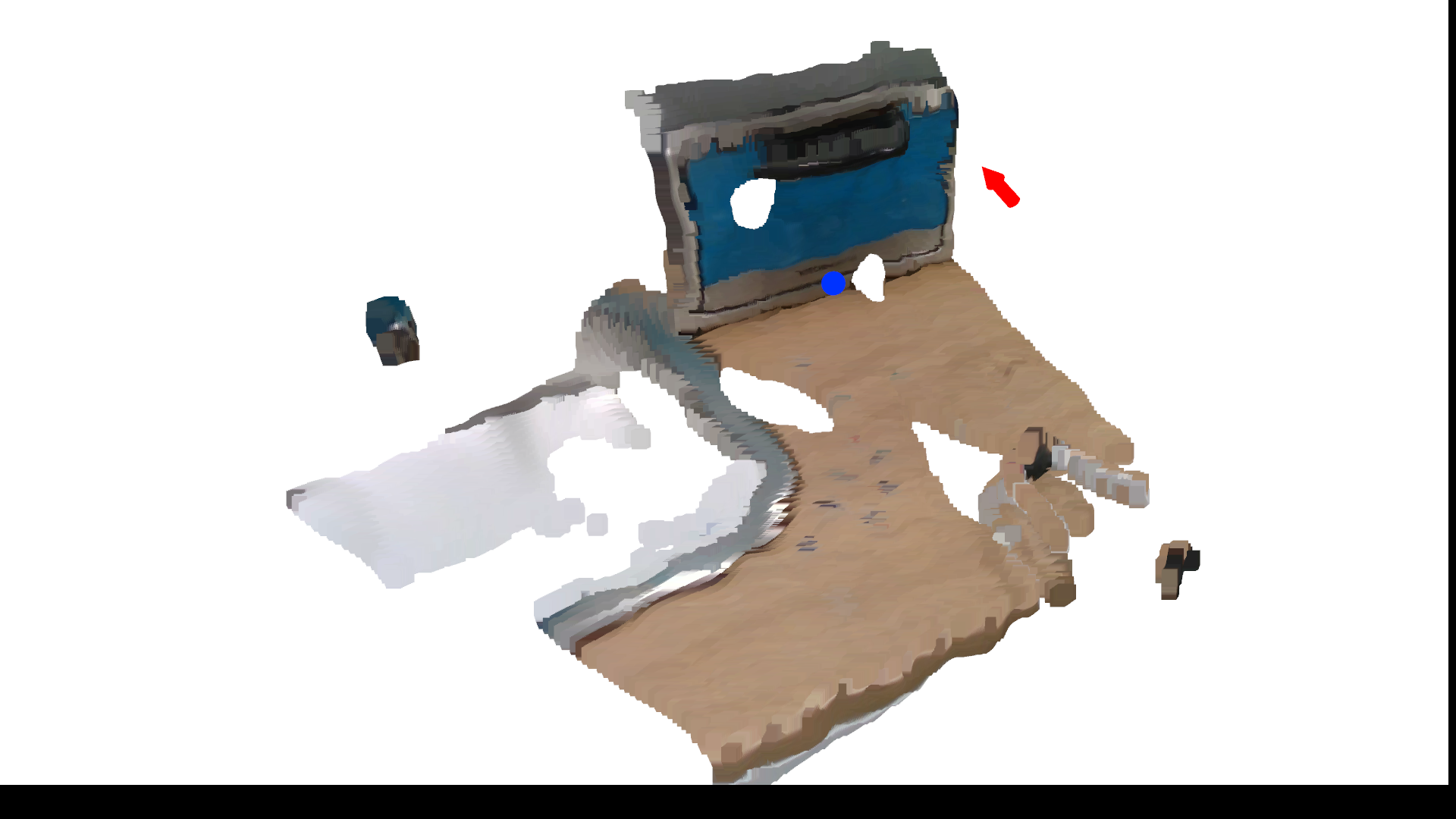}
    }
    \subfigure[]{
        \includegraphics[width=0.15\textwidth,trim=50em 10em 35em 5em, clip]{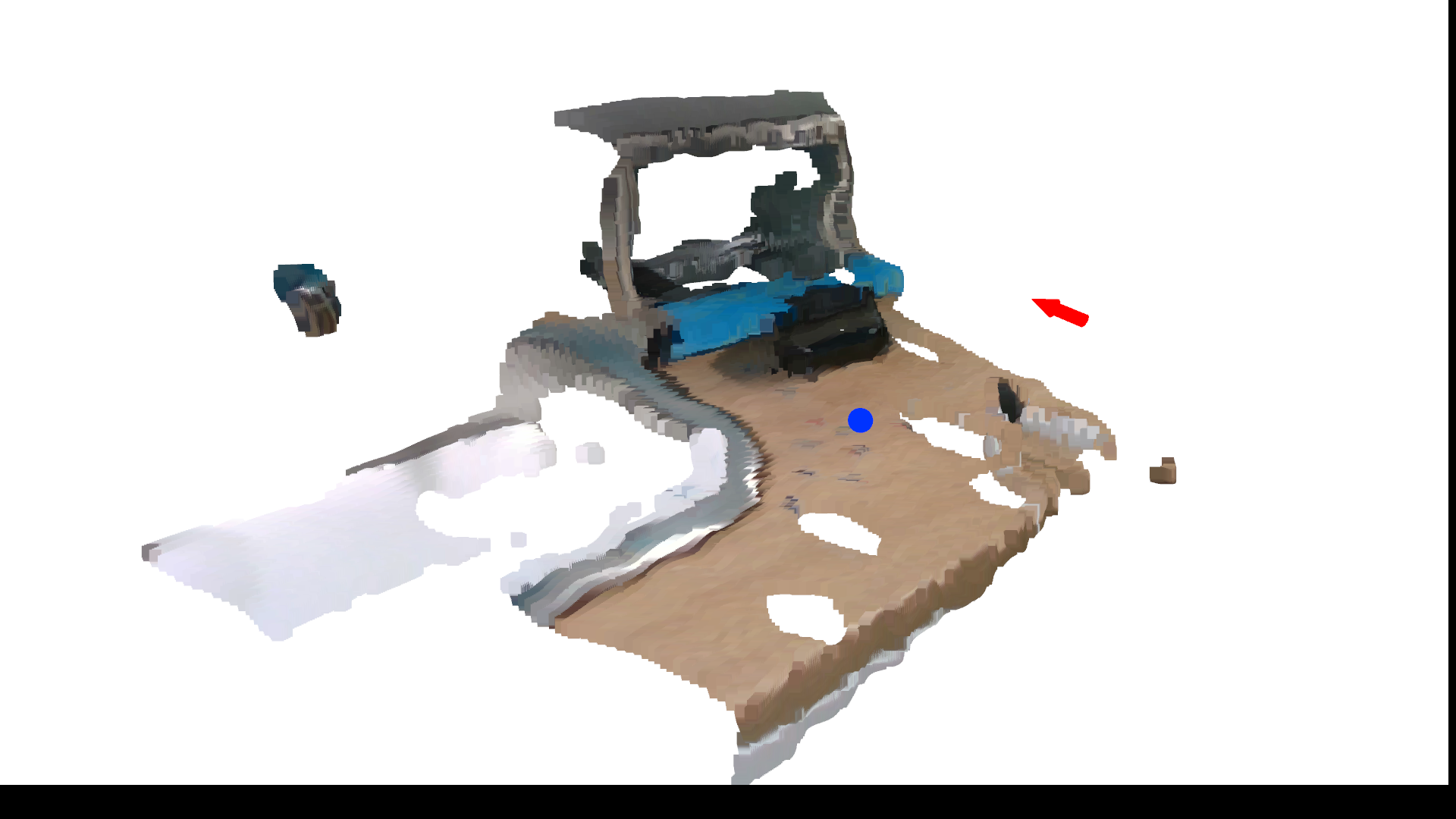}
    }
    \caption{Visualizations of the attention for the MRP stream in three different settings. (a) demonstrates the MRP attending to the bottle. (b) and (c) visualize attention for the MRP by the same \saner{} module to two different configurations of the oven, demonstrating how \model{} attends to the handle of the oven. (d-f) represent the attention for the offset for the same settings. (g-i) represent the MRP with a blue sphere and an arrow for the predicted action.}
    \label{fig:attn_examples}
\end{figure}

This version of attention can be seen as analogous to the spatial attention maps used by CBAM~\citep{woo2018_cbam}, using {set aggregation} modules in place of convolution. The context of each point, which determines its importance, is determined by the structure of points in its neighborhood. For example, the oven is present in every task; however, as we can see in Figure~\ref{fig:attn_examples}, it is attended to less in the presence of a bottle, since a different task is being implicitly indicated in our aggregated embeddings.


\subsubsection{Dual-Stream Architecture} 
\textbf{Most Relevant Point (MRP) Prediction.}
Ideally, finding the most relevant point for a particular task would allow us to focus only on high-level task relevant information, ignoring background objects and distractors. 
To encourage this, we remove points from the point cloud in a rectangular prism surrounding the known location of the end-effector. Due to the fact that when using only a small number of demonstrations the current position of the end-effector is a highly accurate signal for the next position; as discussed in Sec.~\ref{sec:model_ablations}, an MRP predictor without this augmentation tends to overfit to these points. 
This point cloud is then passed into an \mpointnet{} module to compute a weighted average position, {our Most Relevant Point (MRP)}.


\textbf{Action Prediction.}
We then predict actions relative to the MRP. In this case the position of the end-effector is highly relevant to the computation of the next desired action. Therefore, we again remove the points around the end-effector, but now we add a rectangular prism in the position and orientation of the end-effector. This is because the end-effector is often occluded, and here we want to ensure it's available for use. 

As before, this point cloud is then passed into a separate \mpointnet{} module. The resulting $\bar{p}$ and $\bar{m}$ are concatenated, along with the current state of the gripper, then passed into four separate MLPs, which compute the position offset ($\Delta p$), rotation (as a quaternion), and gripper state of the end-effector, as well as a prediction for the fraction of the task completed. The offset is added to the MRP to compute the final end-effector position.



\subsubsection{Training \model{}}
\label{sec:abip_loss}
\textbf{Imitation Loss.} To train the policy, we use the action from the demonstration to supervise the position, gripper state, and completion fraction using a mean-squared error loss. We represent orientation as a quaternion and use $1 - (q_{true} \cdot q_{pred})^2$ as the loss, based on~\citet{Huynh2009MetricsF3}. 

\textbf{Offset Loss.} We additionally compute a loss that drives the computed offset toward 0, to further encourage the MRP to encode information relevant to the interaction: $L_{offset} = ||\bar{p}_{off}||_2$

\textbf{Locality Loss.} 
We encourage generalization with a \textit{locality loss} which encourages relevant points to be close together, defined in Equation \ref{eqn:locality}: 
\vspace{-1em}
\begin{equation}
\label{eqn:locality}
    L_{local} = \sum_{i \in P} a_i ||p_i, \bar{p}_i||_2  
\end{equation}

Since these points must also contain information relevant to the task, this allows for a natural convergence of attention around a salient point in the observed point cloud.
In effect, this allows us to more quickly learn to ignore irrelevant features of the environment, such as walls or the table surface, allowing for improved generalization.

\textbf{Removal of Cloning Losses} While SANE uses the policy and value cloning losses from CLEAR, which help maintain old behavior in the presence of new data, with \saner{} this is unnecessary, as we are already training on the true actions observed from the demonstrations. 

\textbf{Removal of Policy Augmentation} Unlike in SANE, \saner{} trains the policy only on new data ($B_{new}$), not a batch augmented from the replay buffer ($B_{aug}$). This allows the module to specialize the new policy to the new setting as quickly as possible. This is related to policy freezing, discussed in Section \ref{sec:saner_creation}.

\subsection{Critic}
\label{sec:saner_critic}
{SANE uses a reinforcement learning critic to estimate two values, a score and an uncertainty, that are used to determine which module to activate and to detect when drift has occurred. For \saner{}, both of these need to be changed to work in the imitation learning setting.}

\textbf{Critic Architecture} In addition to the \model{} policy network, \saner{} uses a separate \mpointnet{} encoder for the critic, plus a 3 layer MLP with hidden dimension 64 to predict the value and uncertainty. Unlike in SANE, \saner{} does not share weights between the actor and the critic. 
Note that the critic does not utilize a locality loss; we found that earlier tasks might focus too strongly on features that were insufficient signals for future tasks, and the critic's performance would suffer.

\textbf{Activation Score}
 Conceptually, the goal of SANE's activation score is to evaluate how well each module will perform in the current context. When applied to the reinforcement learning setting, the standard critic is a natural choice, as it is already designed for that purpose. However, in the imitation learning setting, the choice is less clear. 

An activation score should: (1) always be greater than 0, to minimize unnecessary module creation at the beginning of training; (2) have a consistent maximum value, for consistency 
between tasks and to make hyperparameter selection easier -- we choose a maximum of about
3; 
and (3) be highly sensitive to policy errors on about the same scale as the hardware and tasks permit, and less sensitive to the difference between large errors in action prediction. 

Our last criteria is both the most important and the least trivial. Say we have a module that succeeds at a particular task with an error in the predicted-end effector position of about 1cm. If updates to the policy that induce an error of 2cm would cause the module to start failing at the task, then the score function should drop considerably. However for the same task, if a different module has a policy error of, say, 18cm, then an increase to 20cm should have little effect on the score; both are clear failures, and should be near 0. 

While there are a number of options for functions that meet this criteria, we found it most straightforward to compute a target score based on the distance between the true action and the predicted action. Specifically, we use: $s_{target} = softplus(\beta)( a || p_{true}, p_{pred} ||_2) + b$, where $p_{true}$ is the true end-effector position, $p_{pred}$ is the current prediction of the policy, and $a$, $b$, and $\beta$ are constants. {We use essentially this same metric for the orientation and gripper pose as well, and sum the results to achieve our final score.} More details on the selection of these constants is described in Appendix \ref{appendix:sane_score}. We then train the critic to predict this target score using an L1 loss.

\textbf{Uncertainty Estimation}
In imitation learning, the policy learns much more quickly than in reinforcement learning, due to the fact that RL relies upon credit assignment instead of direct supervision, causing learning to take orders of magnitude more time. While faster training is generally beneficial, {we found that our prediction of uncertainty tended to lag behind the performance of the policy, particularly at the beginning of a task when the policy is changing rapidly. To resolve this, we introduce a multiplicative factor on the uncertainty, for both the new node and the source node, that increases when drift is detected, and decays as the module is used. This gives the uncertainty estimator more time to converge accurately.} 

{More specifically, we defined our modified uncertainty as $u' = (1+f) * u$, where $u$ is the standard uncertainty predicted by SANE, and $f$ is our uncertainty factor. When a drift event is detected and a node is created, $f$ is incremented by $k_{s}$ for the source node and $k_n$ for the new node. When a node is activated, its $f$ decays according to: $f_{t+n} = f_{t} * \gamma^n$, where $n$ is the number of timesteps seen by the module during activation. }

{In SANE, uncertainty is used for node activation and upper bound estimation. We can therefore define $f_{act}$ and $f_{ub}$ separately, with separate $k$ and $\gamma$ values for each. This is useful because, for example, while we want the source node's uncertainty to be reflected in its upper bound to mitigate unnecessary node creation, we do not need the source node to be more likely activate. Uncertainty is also used for lower bound estimation; however, a sensitive lower bound poses little issue, and we therefore neglect $f$ for this case.}

\textbf{Replay augmentation.}
In reinforcement learning it is not feasible to mine negative examples for one module from the others, because it is challenging to estimate what outcome a policy will have in an environment without actually doing it. However, with imitation learning, mining negative examples is trivial. We leverage this advantage by, at each activation step, giving the active module a 10\% chance to exchange a few samples with each other module.

\textbf{Handling noisy predictions.}
While \model{} is training, the point cloud augmentations, particularly the random downsampling before the first set aggregation module, can result in critic predictions with significant noise. We found that the slow critic SANE uses to smooth training did little to impact this source of noise. We opt to remove the slow critic, and instead average all critic computations over 5 runs of the same observation.

\subsection{Module Creation}
\label{sec:saner_creation}



 The main way that catastrophic forgetting is mitigated in SANE is by the creation of new modules~\citep{powers2022_sane}, which allows one to persist prior behavior, while the other learns the new task. Creation occurs when an active module's critic predicts {an upper bound on} predicted value that is lower than the prediction of its anchor. With imitation learning however, once that has happened, the policy has already changed considerably, and significant forgetting has already occurred. 

We mitigate this issue in \saner{} with a naïve strategy: once a module is cloned, we reset both the policy and the critic to their states at the last time the anchor was updated, which is effectively our last known good state. Furthermore, we freeze the policy, preventing it from training further. 






\subsection{Adapting Existing Baselines}
\label{sec:existing_baselines}


Existing continual-learning methods require a few changes to be applied to the learning-from-demonstration context we want to use for our home robotics setting. In particular, we adapted CLEAR and EWC using same change in loss made for \saner{}: using the {changes to the} loss from Section \ref{sec:abip_loss} in place of V-trace, {with the exception that we do not remove policy augmentation for CLEAR, and it is not applicable for EWC}.
 



\section{Experimental Setup}
We demonstrate continual learning in robotics in a kitchen environment by doing the following 4 tasks: picking up a bottle and putting it in the sink, taking a bottle out of the sink, opening a toaster oven, and closing a toaster oven. For each task we collected and train on only two demonstrations.


We perform three evaluations: first, we demonstrate the generalization ability of \model{} on each task independently, second we ablate our key novel contributions on \model{}, and finally we evaluate our continual learning methods on the tasks trained in sequence.

\textbf{Robotic platform.} We utilize the Hello Robot Stretch\footnote{\url{hello-robot.com/}}~\citep{kemp2021_stretch} in all our experiments, shown in Figure \ref{fig:cover}. The Stretch robot is a ``low cost'' robot, composed of an arm that can independently move vertically and horizontally, a wrist with 3 rotational degrees of freedom, and a pinching gripper. The base is also capable of motion, but for our experiments we keep the robot stationary, leaving this for future work. {The camera used to collect observation data is a RealSense D435 mounted on the head of the robot, and remained stationary during all episodes. Demonstration collection, training, and inference were done off-robot on a dedicated desktop, and communication with the robot was done via ROS.}

\textbf{Task selection.}
The tasks were chosen to highlight a variety of kitchen-relevant sub-skills, under the constraint that there are 5 degrees of freedom in the motion of the arm. The two primary skills exercised are pick-and-place and manipulation of an articulated object. The entirety of our train and test settings can be seen in Figure~\ref{fig:cover}, where the opening and closing of the oven are two tasks using the same settings. Notice that with only small perturbations of the object of interest, we observe significant ability to generalize.

Specifically, picking the bottle from the sink requires a precise approach, as pushing the bottle is risky and may tip the bottle. During our evaluations, we move the bottle in all three dimensions, to evaluate the agent's ability to generalize accurately. The choice of a bottle partially filled with water is convenient, as it allows for tuning the task's sensitivity to imprecise actions.
Picking the bottle from the counter is easier in both these respects, allowing us to use this setting to test more significant displacements in all dimensions, as well as robustness to clutter.  

The toaster oven, by contrast, requires learning the curved trajectory of a closing door. An agent that cannot execute the oven interactions accurately will often get stuck, and will need to retreat and re-grasp, as for example is shown in the Appendix in Figure~\ref{fig:regrasp}. We train on two different ovens, and demonstrate generalization using an unseen oven in differing positions, and in the presence of clutter.

Beyond evaluating these specific sub-skills, the tasks we have chosen provide an efficient test-bed for continual learning specifically: multiple interactions with the same object provides opportunity for forward transfer, while similar observations with conflicting behaviors is an efficient way to elicit catastrophic forgetting. 

We collected two demonstrations for each task, and execute 3 out-of-distribution trials. While training the agent on a task, we sample transitions randomly from both demonstrations. More information can be found in Appendix~\ref{appendix:evaluation}.

\textbf{Collecting demonstrations.} We guided the robot through the trajectory using a controller, 
recording actions at critical points (key points) as in prior work~\citep{shridhar2022perceiver}. 
At each key point, we collect the RGB and depth images, as well as the robot's joint states. The joint states are converted into end-effector position and orientation using forward kinematics. {More detail and an example demonstration are shown in Appendix \ref{appendix:demos}.}

\textbf{Implementation Details} \saner{} is based on an implementation of SANE utilizing IMPALA~\citep{espeholt2018_impala}, as provided by CORA~\citep{powers2022_cora}. Each module has a replay buffer size of 625. Since four modules were created during training, \saner{} uses a total number of 2500 stored frames. {CLEAR and EWC were also implemented using IMPALA. For consistency we set CLEAR's total number of replay frames to be 2500. All other hyperparameters are given in Appendix~\ref{appendix:hyperparameters}.}

\noindent \textbf{EWC} Unfortunately, while we tried training EWC using $\lambda$ in [1000, 10000, 100000], we found no value that worked reasonably in our setting, and opted not to run EWC on the robot. More detail is provided in Appendix \ref{appendix_ewc}. 

\vspace{-1em}
\subsection{Evaluation}
\label{sec:evaluation}

Evaluation was run by randomly selecting one of the two demonstrations, and initializing the robot to its starting configuration. 
We scored performance according to a rubric, with partial task completion earning partial credit. Scoring details are given in Appendix~\ref{appendix:evaluation}. Roughly speaking, the robot earns a higher reward the more of the trajectory it successfully executes.

For \model{} performance and ablations, we present the mean of reward over three trials. For the continual learning results presented in Section~\ref{sec:sequential_results}, we captured performance data before and after training on each task, as well as final performance at the end of each run. 
While it might be preferred to collect data for \textit{all} tasks at the end of each task, or even significantly more frequently as per~\citet{kirkpatrick2016_ewc},
this is infeasible due to the time-consuming nature of real-robot experiments.
Additionally, we never test in comparison with a randomly-initialized policy as in \citet{chaundhry2018_intransigence}, to avoid damage to the robot or environment.


\textbf{Metrics.}
We use the following metrics, based upon existing methods~\citep{chaundhry2018_intransigence, powers2022_cora}, modified to capture the most information using the fewest number of interactions with the environment.

We use the notation that $R_{x, y}$ indicates average performance on task x after training on task y, where $y=\varoslash$ indicates before training on anything, and $N$ indicates the index of the final task. $R_x$ indicates the average performance on a task trained alone (non-sequentially). We assume the reward of a random policy, $R_{i, \varoslash}$, is approximately 0, which will hold in general for complex tasks with a short time horizon. We additionally assume tasks are only seen once, not cyclically.

\vspace{-1em}
\begin{itemize}
\itemsep0em 
    \item Final performance $R_{final, i} = R_{i, N}$: reward after training is completed on all examples.
    \item Performance improvement $\Delta R_i = R_{i, i} - R_{i, i-1}$: change in reward/score by training on the task.
    \item Zero-shot forward transfer $ZSFT_i = R_{i, i-1} - R_{i, \varoslash} \approx R_{i, i-1}$: reward achieved at the very start of a task, before any explicit training on it is done.
    \item Cumulative forgetting $F_i = R_{i, i} - R_{i, N}$: decline in reward after learning new skills.
    \item Intransigence $I_i = R_{i} - R_{i, i}$: relative inability to learn compared to a stand-alone model
\end{itemize}

The first four metrics only require that we evaluate a policy's performance on two tasks at each task transition -- the task just trained, and the task about to be trained -- and on all tasks once at the end. The fifth metric, intransigence, additionally requires that we train and evaluate a separate model per task. This simplification from the standard, of evaluating across all tasks, allows our evaluation to scale linearly in the number of tasks, instead of quadratically.

Zero-shot forward transfer, performance improvement, and cumulative forgetting together provide us with an overall view of the lifetime of performance for each task: ZSFT indicates how much of a direct boost prior tasks provided, $\Delta R$ indicates direct improvement from training on the task, and $F$ tells us how much is lost via the remainder of the tasks. Finally, where ZSFT tells us of the immediate impact prior tasks had, intransigence tells us about the latent effects, e.g. via decreased capacity or ineffective representation.

To make at-a-glance comparison easier, we present $F$ and $I$ as $-F$ and $-I$ instead; this is because the rest of the metrics indicate better performance the larger they are, whereas these two, by convention, are worse. We refer to $-F$ as recall, and $-I$ as plasticity.

\vspace{-0.5em}
\subsection{Evaluating Attention-Based Interaction Policies}
First, we examined the ability of \model{} to generalize to out-of-distribution variations of each task. 
We trained on each task separately using CLEAR~\citep{rolnick2018_clear}, and then evaluated on each of the training settings, as well as in the out-of-distribution generalization settings visualized in Figure~\ref{fig:cover}. Results are shown in Table \ref{tab:clear_single}.

Our specific goal is to demonstrate that this model is sufficiently capable of learning skills in the few-shot setting, in order to be useful as a building block of a continual learning system.
For this to be the case:
(1) \model{} must learn well enough that catastrophic forgetting would be observable, (2)  the domain must be challenging enough for forward transfer to be meaningful, and (3) general representations and policies must be attainable, so that the skills learned are practical.

We see generally high performance across both in-distribution and out-of-distribution settings, with a few exceptions. {While performance is not directly comparable, we see that our method compares favorably to similar settings presented by other work using the Stretch in the home environment~\citep{bahl2022humantorobot, parashar2023spatiallanguage, pari2021_visualimitatation}} As discussed above, \model{}'s performance on each task highlights different strengths and weaknesses of the method. The method performs well on the \textit{Bottle To Sink}  task, particularly on the unseen settings, which indicates that the method is robust to significant displacements of the target object, and to distractors -- key features of our MRP-based method. It suffers a bit on the \textit{Bottle From Sink} generalization task, however. This task is highly sensitive to small errors in grasp trajectory, which indicates that while the MRP is capable of identifying our object, the action prediction is an area of improvement.



Overall, it is clear that \model{} can capture our task, and that the task is challenging enough to be interesting.
Additionally, we specifically demonstrated both that the method is capable of generalizing to objects in unseen, out-of-distribution positions, as well as to unseen objects. Our policies were able to adapt to significant shifts in position and scene composition, including being able to open and close an unseen oven. 


\begin{table}[bt]
    \centering
    \tiny{
        \begin{tabular}{c c c c}
        \toprule
        Task & In Distribution: Demo 0 & In Distribution: Demo 1 & Out of Distribution \\
        \midrule
        Bottle To Sink   & $0.80 \pm 0.34$ & $0.87 \pm 0.23$ & $1.0 \pm 0.0$ \\
        Bottle From Sink & $0.60 \pm 0.40$ & $1.0 \pm 0.0$   & $0.47 \pm 0.23$\\
        Open Oven        & $0.20 \pm 0.20$ & $0.80 \pm 0.20$ & $0.87 \pm 0.11$\\
        Close Oven       & $0.53 \pm 0.50$ & $0.33 \pm 0.12$ & $1.0 \pm 0.0$ \\
        \midrule
        \textbf{Average} & 0.53 & 0.75 & 0.83 \\
        \bottomrule
    \end{tabular}
    }
    \caption{
    Performance of CLEAR~\citep{rolnick2018_clear} on ``in distribution'' and ``out of distribution'' tasks, using \model{} as the underlying policy representation.  We observe significant generalization ability.  
    }
    \label{tab:clear_single}
\end{table}


\subsection{\model{} Ablations}
\label{sec:model_ablations}

Having demonstrated that \model{} is a capable building block for continual learning, we proceed to analyze the effects of the design decisions made in its creation. In particular, we ablate by comparing the \model{} to methods that: 1) remove the locality loss for the MRP, 2) remove the locality loss for the offset, 3) use a single-stream variant of the model where the MRP is predicted in the same stream as the offset, 4) use a version with no MRP at all. Results are shown in Table \ref{tab:ablations}. All results are provided as the average over three generalization trials. 

\model{} outperforms the ablations significantly, with three of the four failing to achieve anything better than a partial grasp. 
 When trained without the locality loss on the offset-prediction head, by contrast, the model succeeded in completing two full trajectories, but failed entirely at the third. However, the case where it failed is informative: it was the setting with distractor objects. 

In summary, \model{} has qualitatively better representations for generalization, finds the object of interest more reliably, and is more capable of executing trajectories to completion.

\begin{table}[bt]
    \centering
    \small{
    \begin{tblr}{Q[c,m]Q[c,m]Q[c,m]Q[c,m]Q[c,m]Q[c,m]}
        & \model{}  & No MRP Locality & No Offset Locality & Single Stream & No MRP \\
        \hline \\
       Bottle to Sink & $1.0 \pm 0.0$ & $0.0 \pm 0.0$ & $0.67 \pm 0.57$  &  $0.0 \pm 0.0$    &  $0.067 \pm 0.12$     \\
    \end{tblr}
    }
    \caption{Ablation experiments on the generalization (out-of-distribution) setting. \model{} provides better generalization from very few examples, including robustness to unseen, out-of-distribution object poses and new objects.
    }
    \label{tab:ablations}
\end{table}


\subsection{Sequential Tasks}
\label{sec:sequential_results}

\begin{figure}
    \centering
    \includegraphics[width=0.18\textwidth,trim=40em 5em 60em 40em, clip]{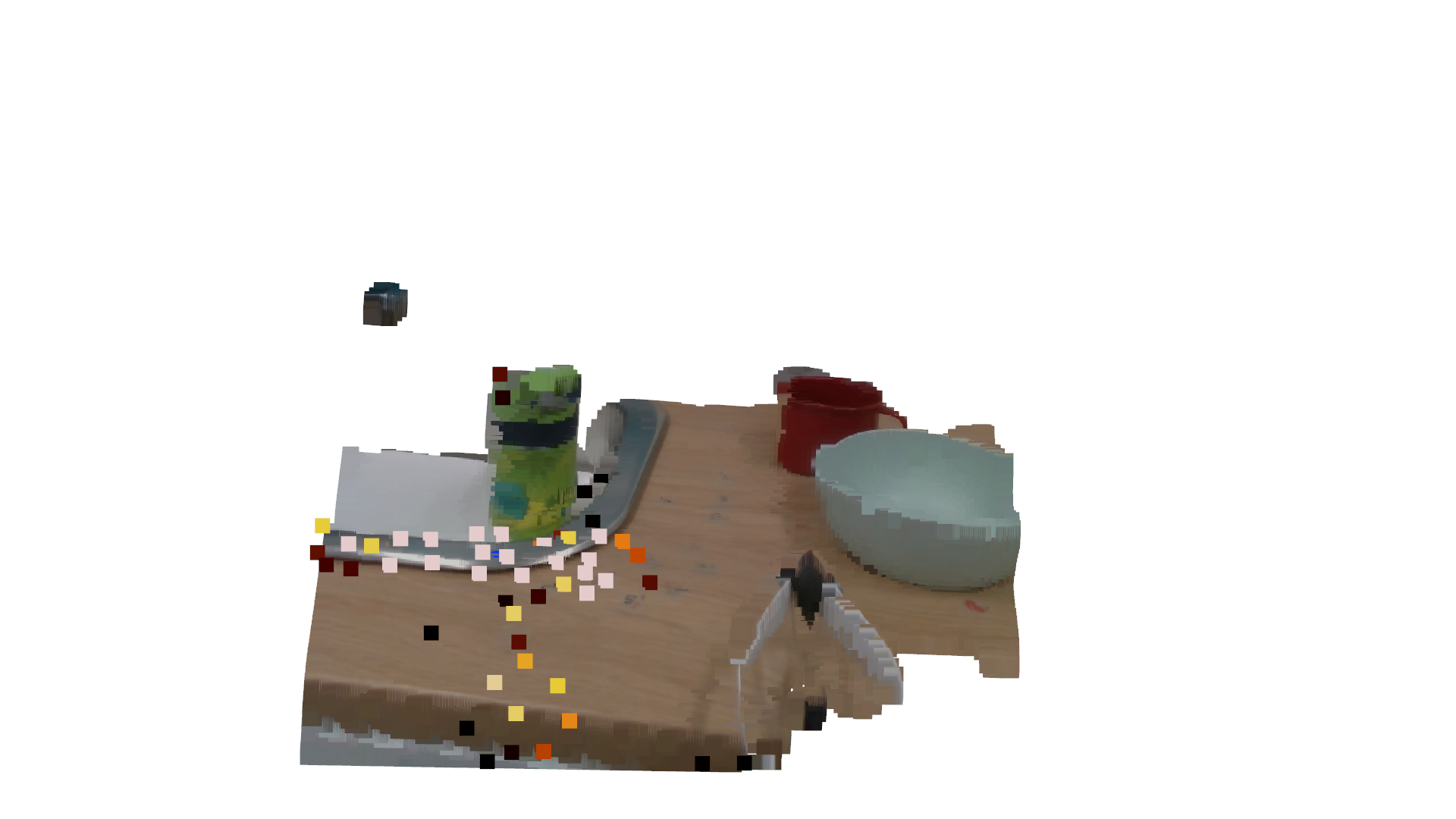}
    \includegraphics[width=0.18\textwidth,trim=40em 5em 60em 40em, clip]{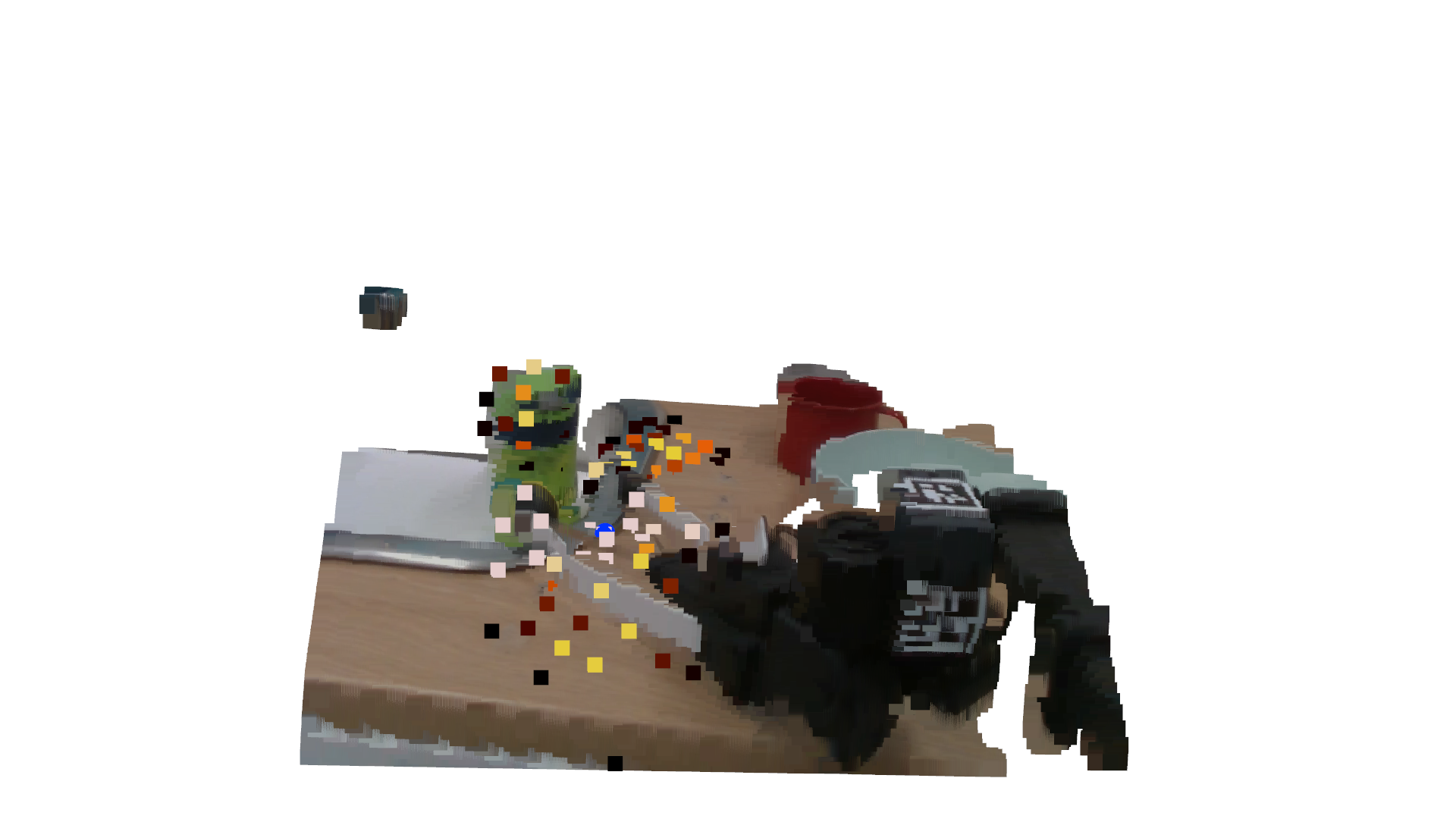}
   \includegraphics[width=0.18\textwidth,trim=40em 5em 60em 40em, clip]{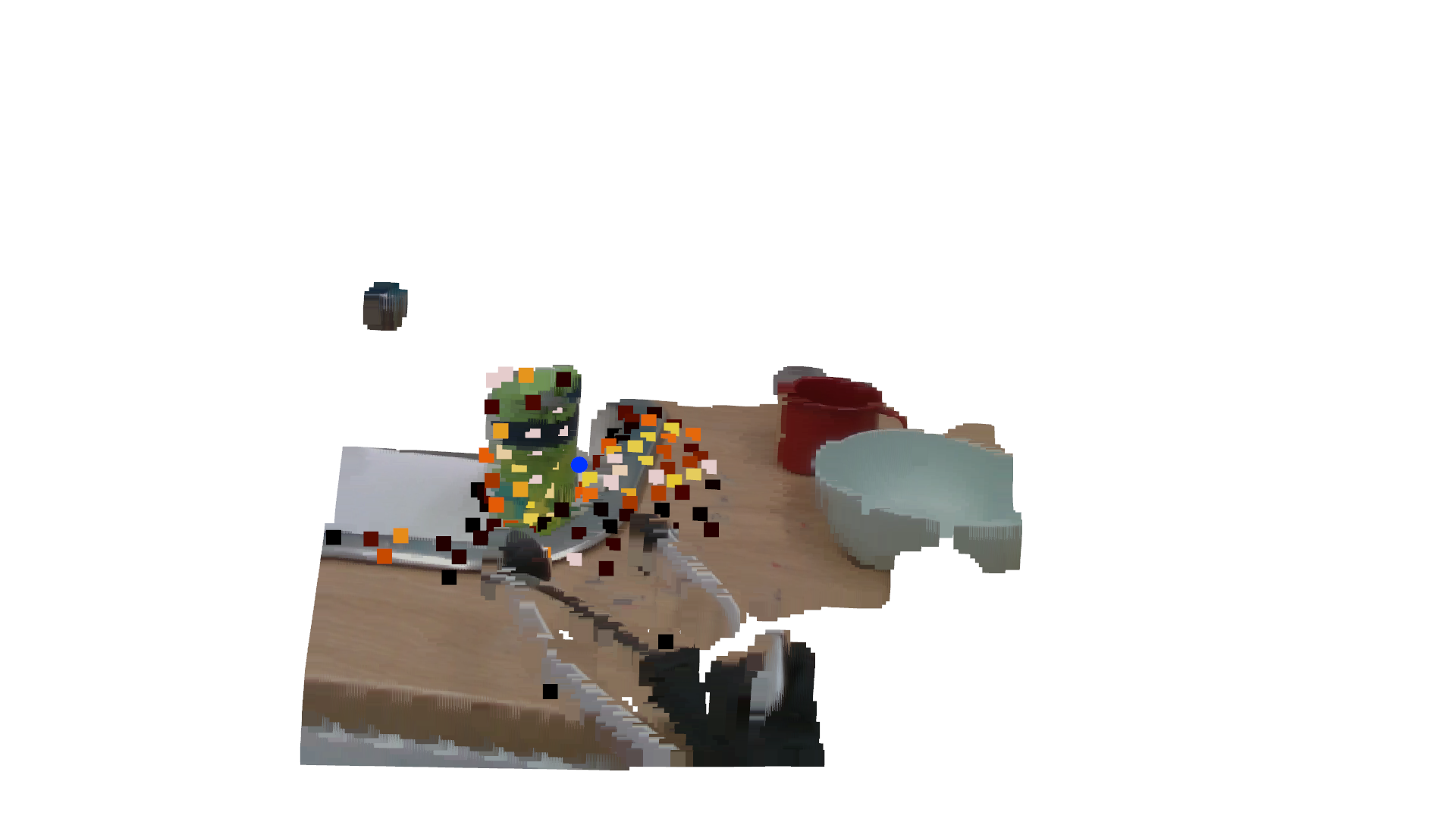}
    \includegraphics[width=0.18\textwidth,trim=40em 5em 60em 40em, clip]{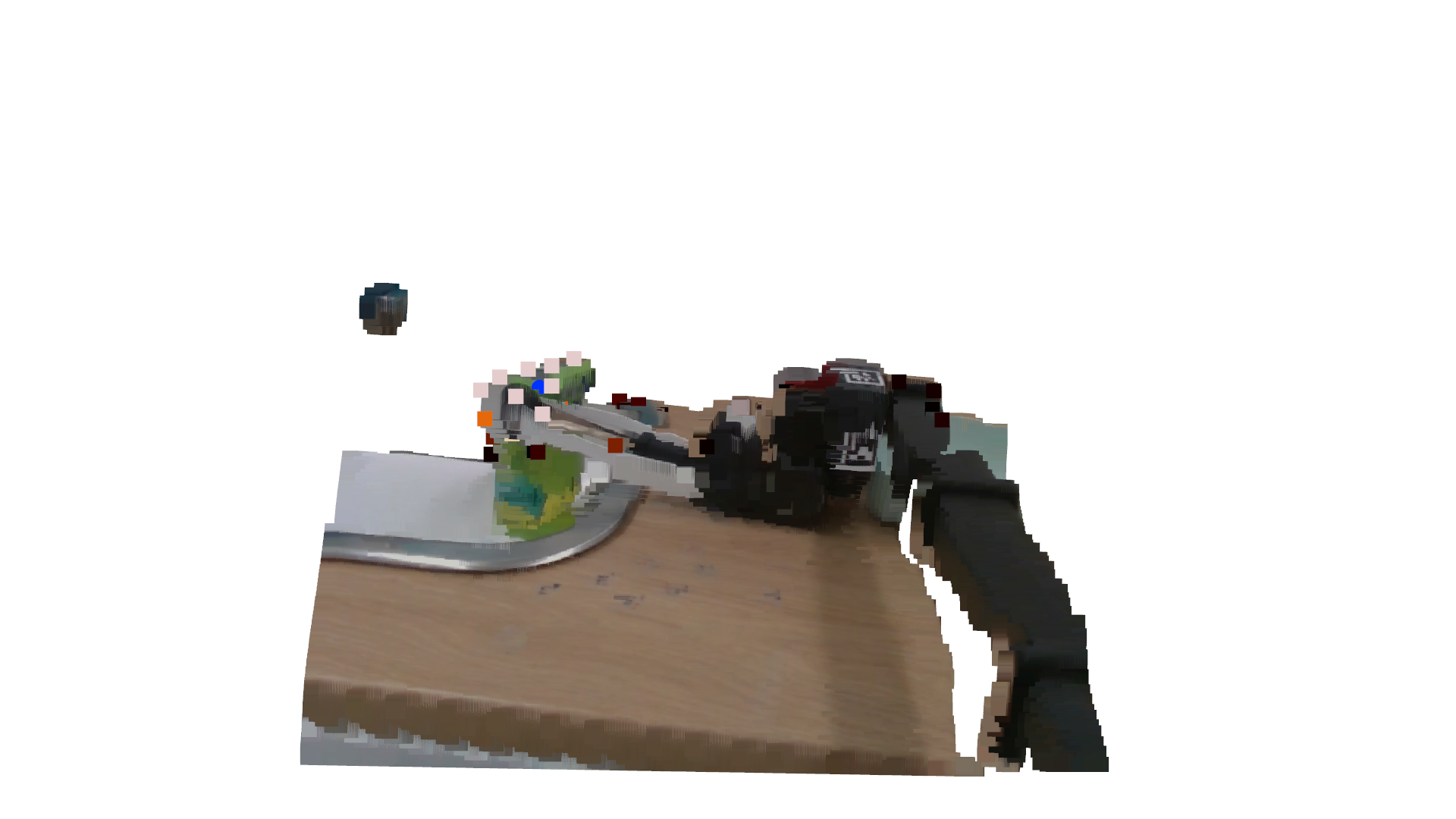}
    \caption{Visualization of the attention for the MRP for \saner{} as it evolves over the episode, showing convergence to the object of interest. The MRP is represented, where visible, with a blue sphere.}
    \label{fig:mrp_evolution}
\end{figure}

We compared the performance of \saner{} and CLEAR in our continual learning setting, where we train sequentially on 4 tasks. We evaluate only on the generalization settings; results are given in Table \ref{tab:seq_learning}. 
%
While CLEAR tends to learn more via direct task training ($\Delta R$), it also exhibited significant forgetting, particularly on the bottle tasks
By contrast, \saner{} uniformly maintains what it has learned, even exhibiting increased performance on the ``Open Oven'' task after training on later tasks.

Additionally, whereas CLEAR observed \textit{some} zero-shot forward transfer, \saner{} experienced a considerable amount, on the ``Close Oven'' task in particular. This likely contributes to \saner{}'s lower $\Delta R$ score, as it was largely capable of solving the task prior to training on it. 
Due to \saner{}'s ensemble nature, it is likely that the separation of representations enabled a more seamless transfer than was possible with CLEAR's single network.

Qualitatively, CLEAR's behavior was somewhat erratic, tending to first approach every object as if it were an oven. In one instance, it executed a seamless oven opening trajectory against the side of a toaster, and in another it grabbed the handle of the bottle and pulled it down in a similar way.
\saner{}'s behavior, while imperfect, is more consistent and interpretable. Since \saner{} modularizes its behaviors, one can roughly know what \saner{} will do by observing which module it activated. However, the generally lower performance as compared to the single task setting indicates significant room for improvement. 
We show an example failure in Fig.~\ref{fig:mrp_evolution}. Its attention shifts first to the edge of the sink, then to the bottle itself as it attempts a grasp. The grasp, however, fails, due to inaccurate timing of gripper closure. Similar grasp-timing related issues were seen in many of the failures in other tasks as well.

The issue may come down to the fact that CLEAR trains on 
effectively seven times as much data as \saner{}.
\saner{} only trains the active module's actor on the current batch of data, instead of an augmented batch, as \saner{}'s critics and CLEAR both do.
As such, we see that after training the first task, \saner{}'s loss is twice CLEAR's, giving further support to this hypothesis. We will investigate solutions in future work.  
%
%
Overall, however, \saner{} has demonstrated utility in the robotics setting, by being able to learn general behavior from only two examples, with essentially no forgetting and a non-trivial amount of forward transfer.

\begin{table}[bt]
    \centering
    \tiny
    \begin{tabular}{cl | ccccc | ccccc}
    \toprule
         &         \textbf{Single Task} &       \multicolumn{5}{c}{\textbf{CLEAR}}  & \multicolumn{5}{c}{\textbf{\saner{}}} 
        \\
        & $R_{i}$ & $R_{final}$ &  $\Delta R$ &  ZSFT   & -F & -I & $R_{final}$ &  $\Delta R$ &  ZSFT   & -F & -I  \\ 
        \hline
        Bottle To Sink &   $1.0$    &   $0.067$       &    0.87         &    ---     &  -0.80  &      -0.13  &  $0.33$    &    0.33         &   ---      & 0  &  -0.67  \\
        
        Bottle From Sink &  $0.47$   & $0.13$       &     0.067        &    0.20     & -0.13  &      -0.33   &  $0.13$   &    -0.067         &    0.13 &  0.067 &  -0.40  \\
        
        Open Oven     &    $0.87$    &   {0.33} 
        &      0.20       &    0.0     &  {0.13} 
        &        -0.60   &  {0.33} 
        &      0.33       &     0.0      &  {0.0}
        &  -0.53  \\
        
        Close Oven    &     $1.0$   &   $0.87$         &     0.73        &    0.13     & 0.0  &       0.13   &  $0.87$   &       0.20      &     0.67   & 0.0  &  -0.13  \\

        \hline
        \textbf{Average}  &     0.84   &   {0.35}         &     \textbf{0.47}      & 0.11        &    {-0.20}     & \textbf{-0.30}   &  {\textbf{0.42}}   &       0.20      &     \textbf{0.26}      & {\textbf{0.017}}   & -0.44    \\
\bottomrule
    \end{tabular}
    \caption{Comparison of two methods for continual learning: CLEAR and \saner{}, as demonstrated on our setting given only a handful of demonstrations, across a number of different metrics. We have negated Forgetting (F) and Intransigence (I) so that in all cases a larger number is preferred. 
    }
    \label{tab:seq_learning}
\end{table}

\vspace{-1em}
\section{Conclusions}

In order to enable robots to operate in the home setting, they need to be able to learn continually from small amounts of data. 
To this end, we proposed \saner{}, an ensemble method adapted to the robotics setting.
We demonstrated, on a set of 4 kitchen skills, utilizing a Stretch robot, that it is capable of learning new skills and generalizing to unseen settings with forward transfer and while mitigating the effects of catastrophic forgetting, out-performing a strong baseline on these metrics.
\saner{} is built on top of \model{}, which can function as a simple building block capable of learning highly useful, very generalizable interaction policies. Finally, we presented the fundamentals of our environment design and a set of continual learning metrics simplified to be viable for the robotics case.
In conclusion, we demonstrate how we can deploy continuous learning techniques like SANE~\citep{powers2022_sane} and CLEAR~\citep{rolnick2018_clear} to a limited-data robotics context.

\section{Acknowledgements}
We'd like to thank Priyam Parashar and Austin Wang for their support in this work, as well as the Hello Robot team for their help in working with the Stretch robot.

\newpage

\bibliography{references}
\bibliographystyle{collas2023_conference}

\newpage
\section{Appendix}
{\subsection{EWC Comparison: Performance on Demonstration Data}}
\label{appendix_ewc}

\begin{figure}[H]
    \centering
    \includegraphics[width=0.2\textwidth, trim={0 0 18em 0},clip]{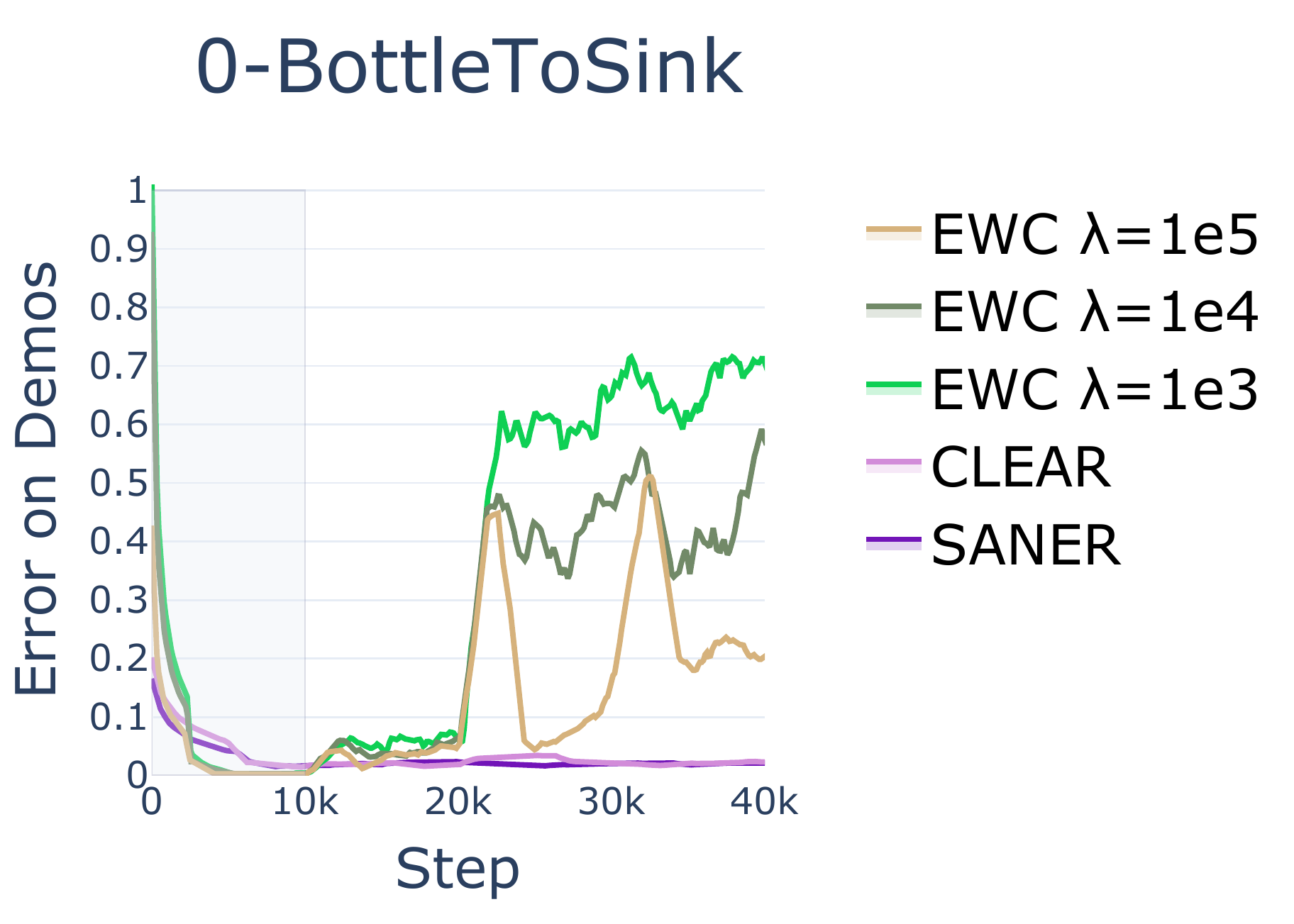}
    \includegraphics[width=0.2\textwidth, trim={0 0 18em 0},clip]{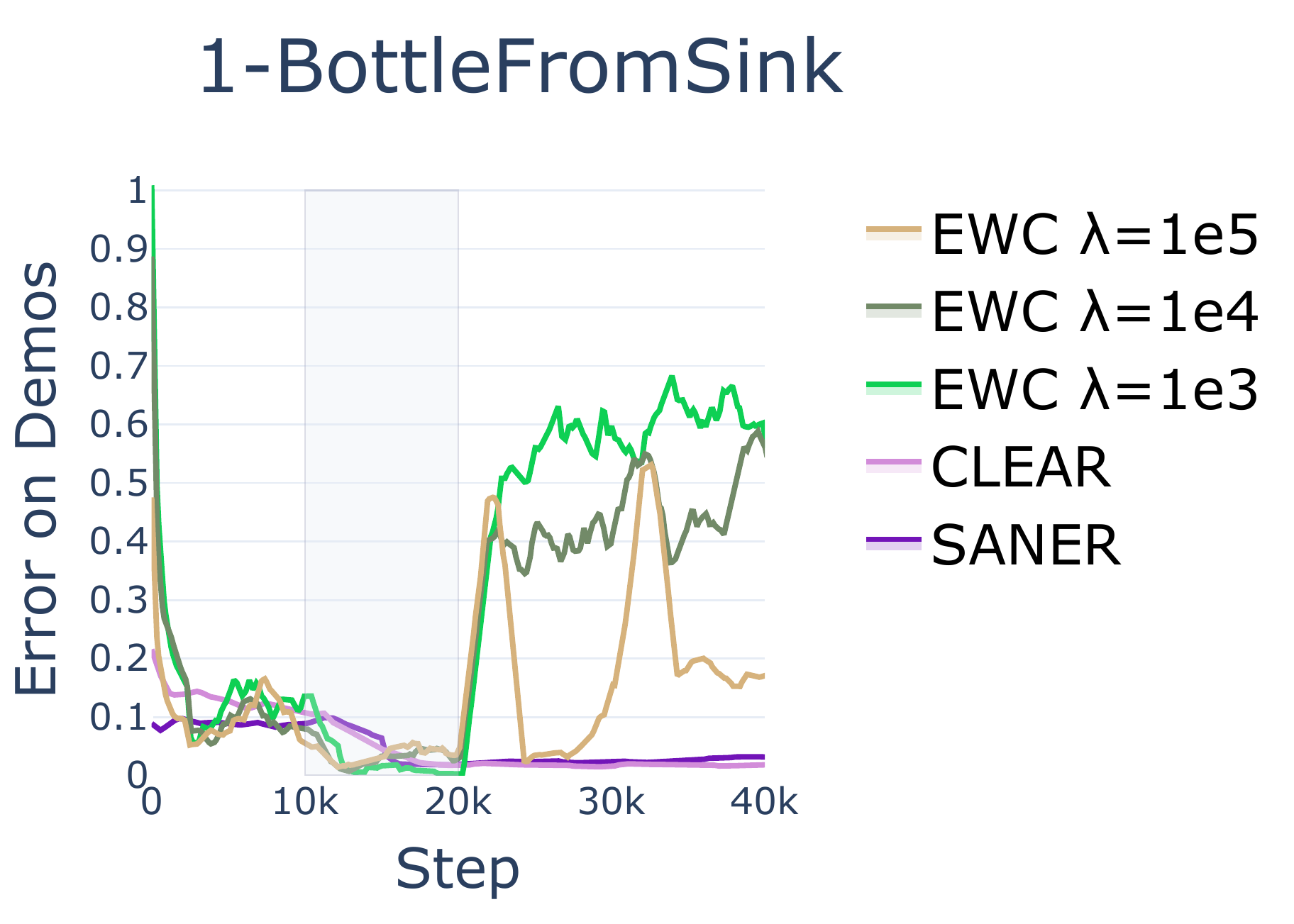}
    \includegraphics[width=0.2\textwidth, trim={0 0 18em 0},clip]{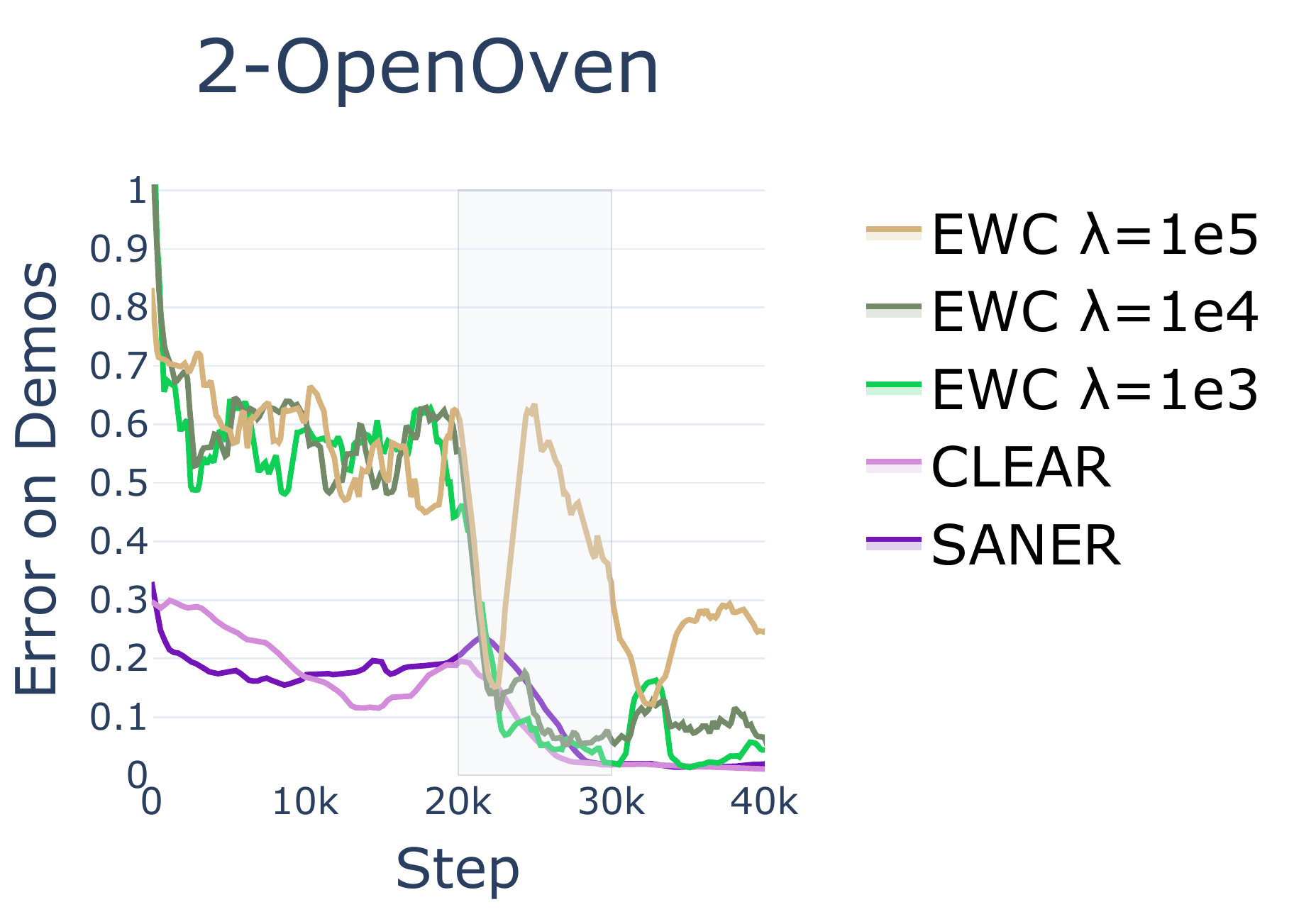}
    \includegraphics[width=0.3\textwidth, trim={0 0 0em 0},clip]{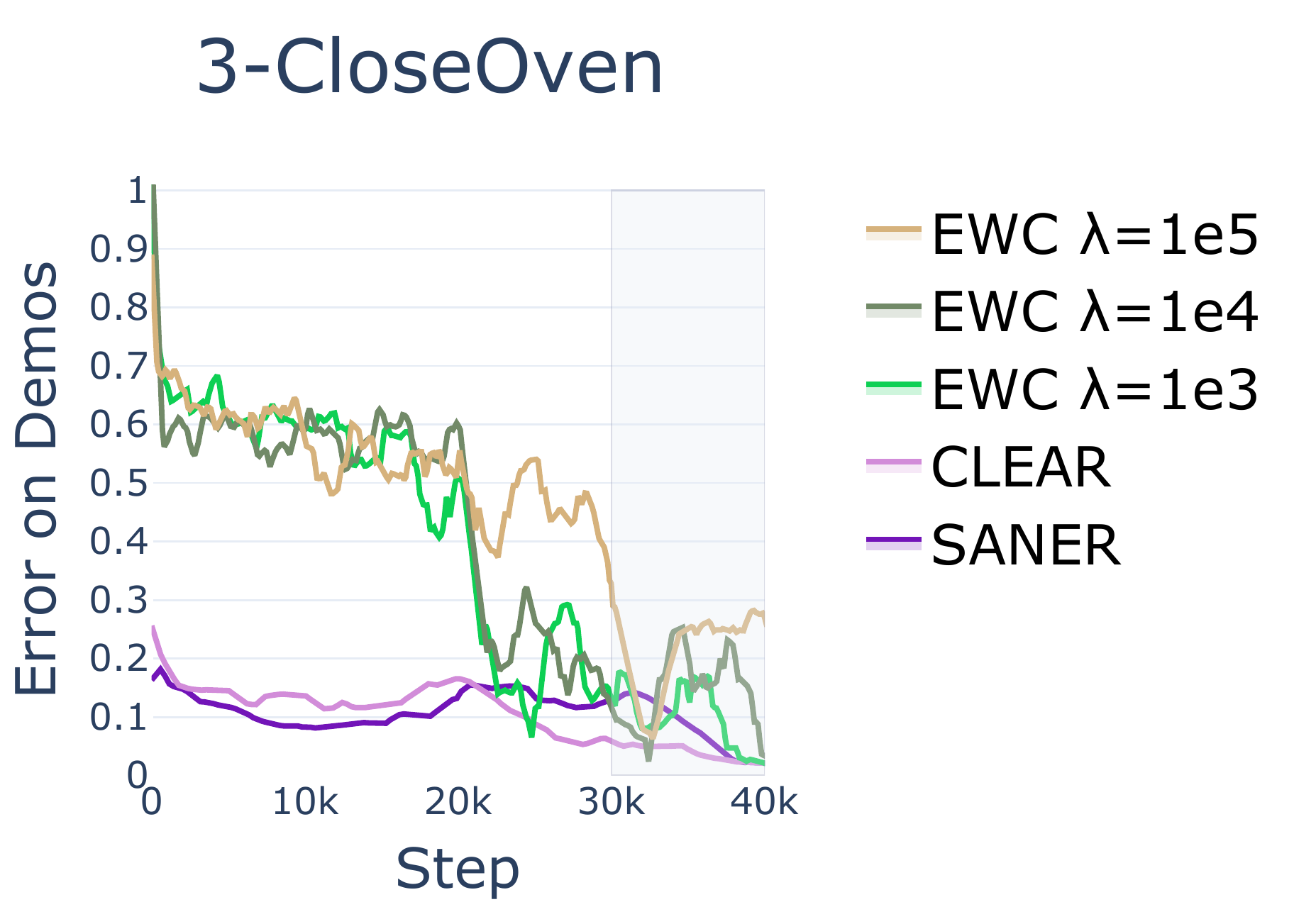}
    \caption{Continual evaluation of the error on demonstrations for each of the methods.}
    \label{fig:demo_eval}
\end{figure}

{
Results from evaluating the methods on the demonstration data are shown in Figure~\ref{fig:demo_eval}. The grey box indicates the region where each task was trained. Reported error is the same as the loss used to train the policy. 
}

{
Note that at the lower $\lambda$ values for EWC, we observe significant forgetting of the earlier tasks, but at the highest value, latter tasks struggle to learn in the presence of the regularizing EWC loss (triggered after 2k steps of each task). We believe that this indicates that the method is not able to find a single parameter set that is able to solve both tasks. Since error was significant, we opted to not risk damage to the robot or environment by running this policy.
}

\subsection{Complexity Analysis of \saner{} vs CLEAR}

{Here we analyze the difference in run-time and memory complexity for SANER vs CLEAR.}

{\textbf{Space Complexity} We hold the total replay buffer size constant for \saner{} and CLEAR: in the experiments presented in the paper, \saner{} uses 625 frames for each of 4 modules and CLEAR uses a total of 2500 frames. In longer sets of tasks with more module creation, we would scale as necessary to maintain consistency. Additionally, while we have not found the size of the networks themselves to be a limiting factor, since \saner{} only activates one module at a time, it would be possible to off-load unused resources as necessary.}

{\textbf{Run-time Complexity} \saner{} trains both the actor and the critic for the active module, in contrast to CLEAR which only trains the actor. However, the \saner{} critic is single-stream (no MRP prediction). In practice, \saner{} takes a bit less than 2x the time to train as compared to CLEAR.} 

{\saner{} also runs the critic for every node every 150 time steps during training, to determine which module to activate. There is a trade-off here between activating infrequently (for speed) and how quickly we can detect and react to drift. During evaluation, we run the critic for all nodes, for each episode. Time for activation is the only run-time factor that scales in the number of modules; at the activation and continual evaluation frequencies we selected, however, this is not a dominant factor in the required time to train.}

{In total, running the 40k timesteps for training these tasks took about 5 hours for CLEAR, and 9.5 hours for SANER.}

{
\subsection{Demonstration Details}}
\label{appendix:demos}
\begin{figure}[H]
    \centering
    \includegraphics[width=0.18\textwidth]{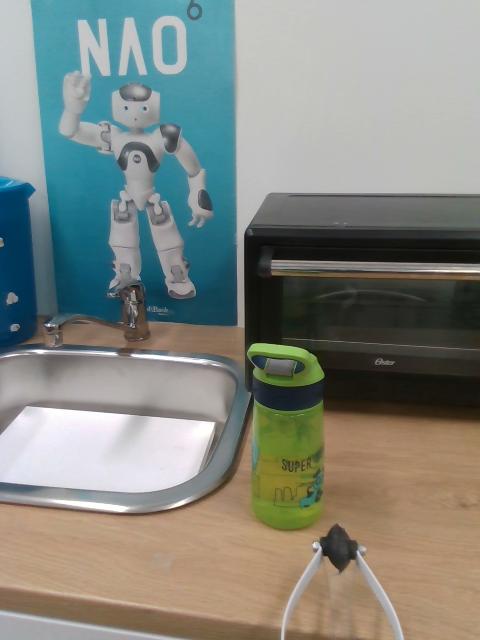} 
    \includegraphics[width=0.18\textwidth]{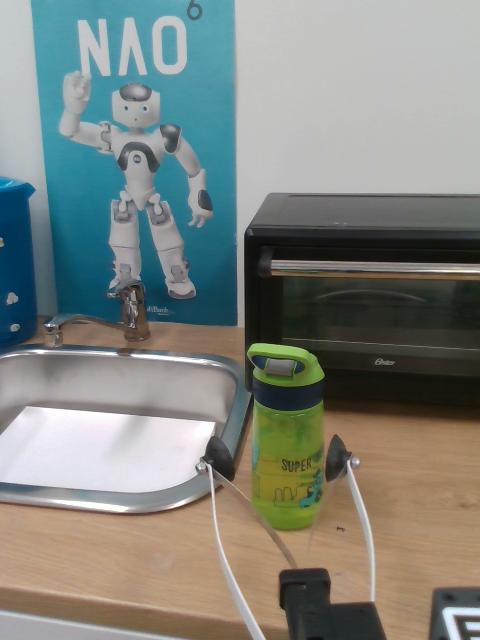} 
    \includegraphics[width=0.18\textwidth]{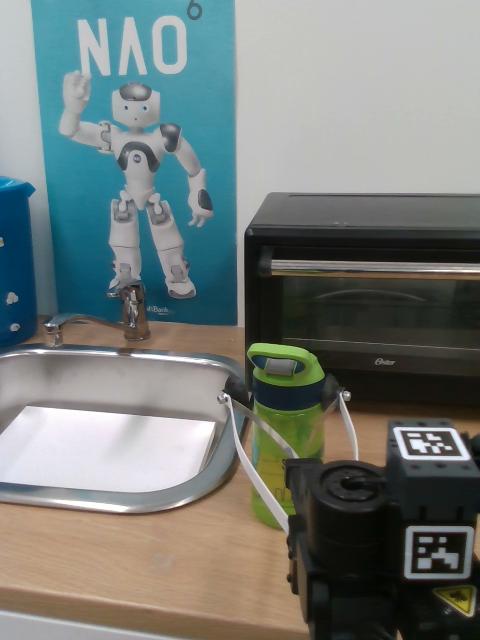}
    \includegraphics[width=0.18\textwidth]{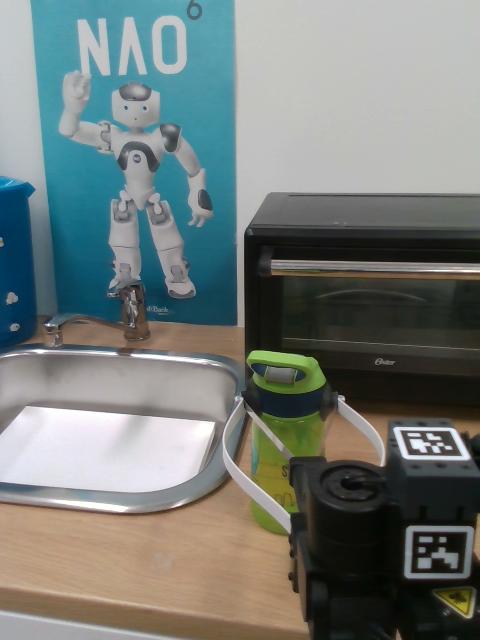} 
    \includegraphics[width=0.18\textwidth]{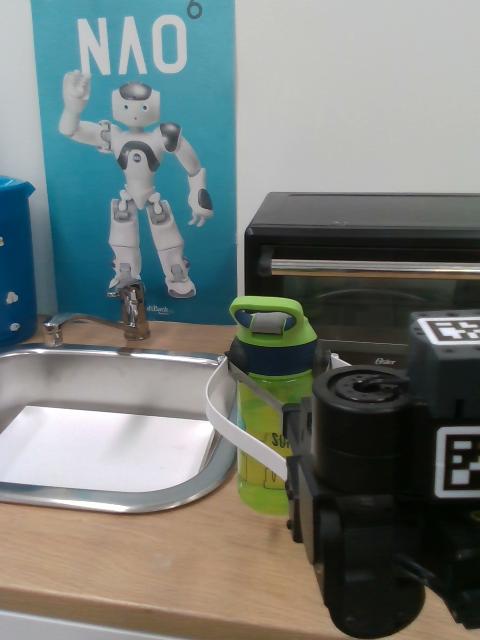} \\ 
    \includegraphics[width=0.18\textwidth]{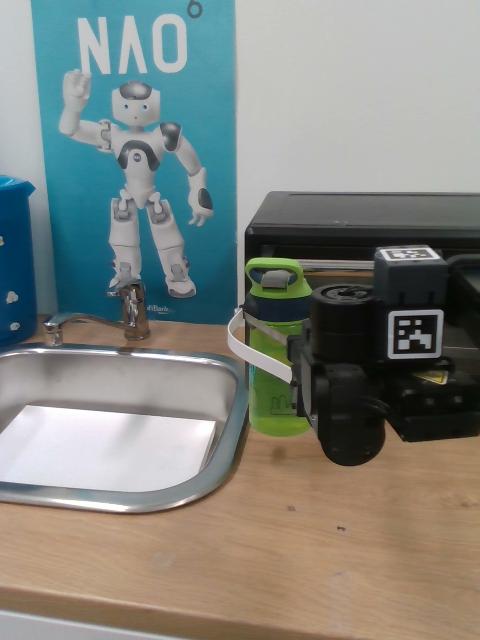} 
    \includegraphics[width=0.18\textwidth]{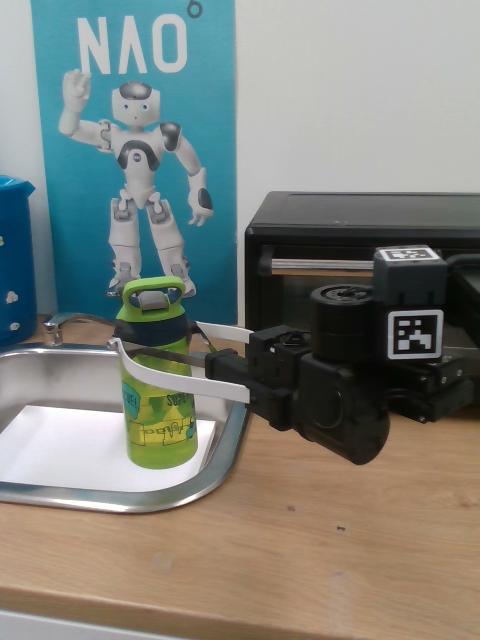} 
    \includegraphics[width=0.18\textwidth]{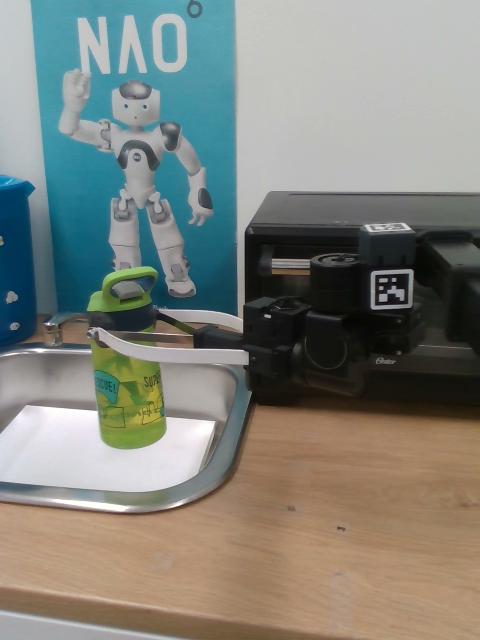}
    \includegraphics[width=0.18\textwidth]{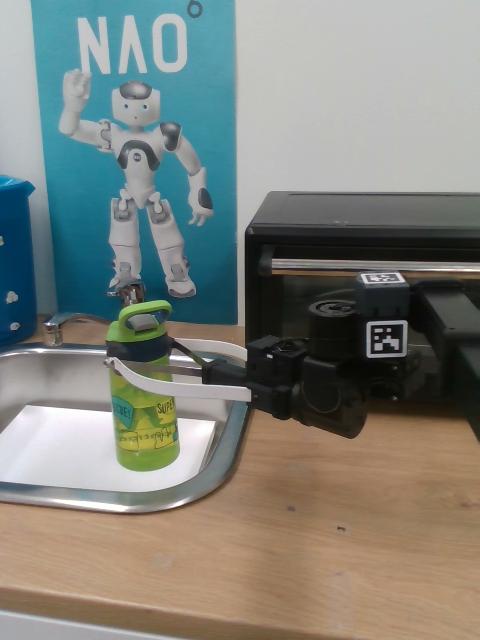} 
    \includegraphics[width=0.18\textwidth]{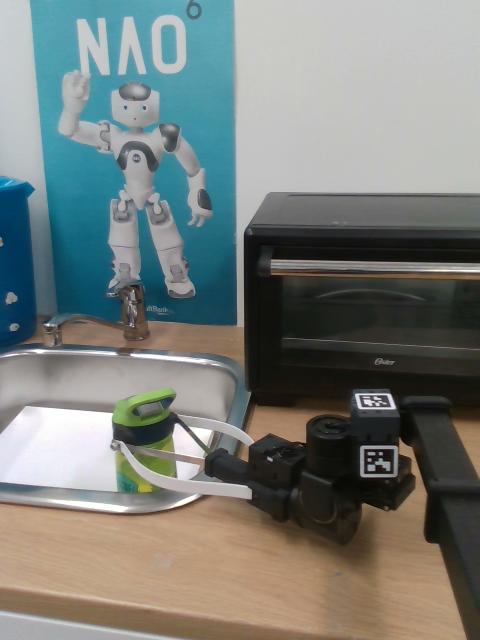} \\ 
    \includegraphics[width=0.18\textwidth]{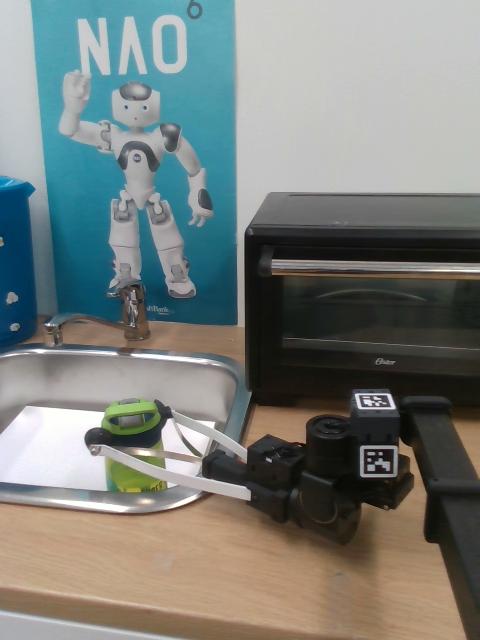} 
    \includegraphics[width=0.18\textwidth]{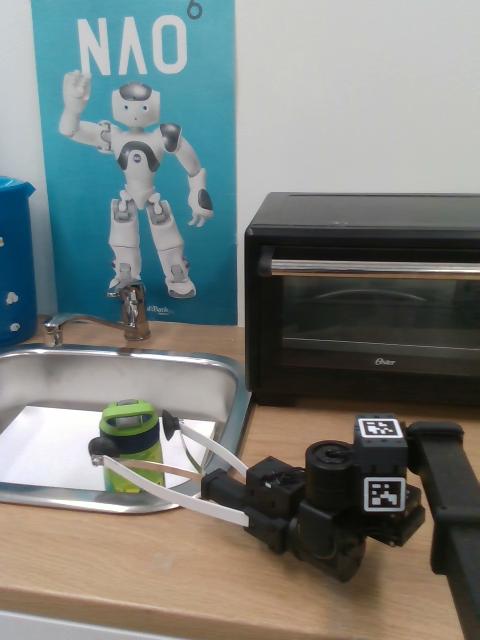} 
    \includegraphics[width=0.18\textwidth]{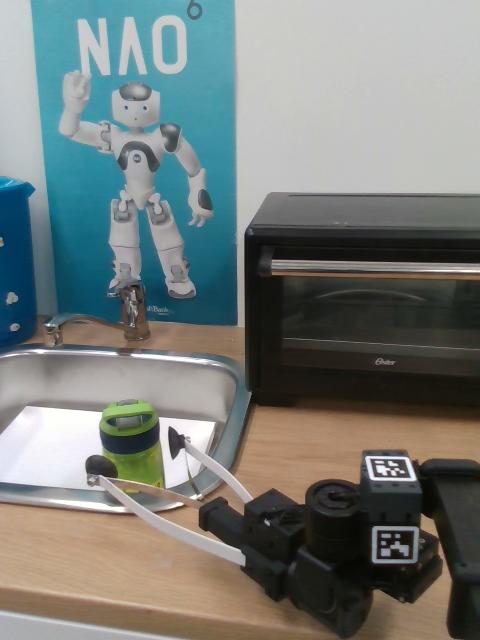}
    \includegraphics[width=0.18\textwidth]{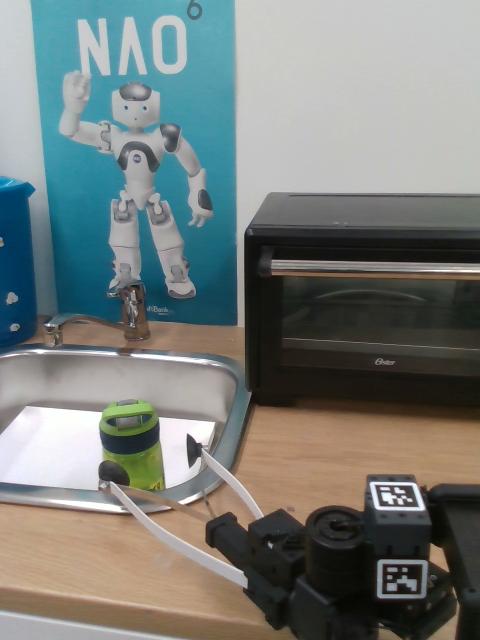} 
    \includegraphics[width=0.18\textwidth]{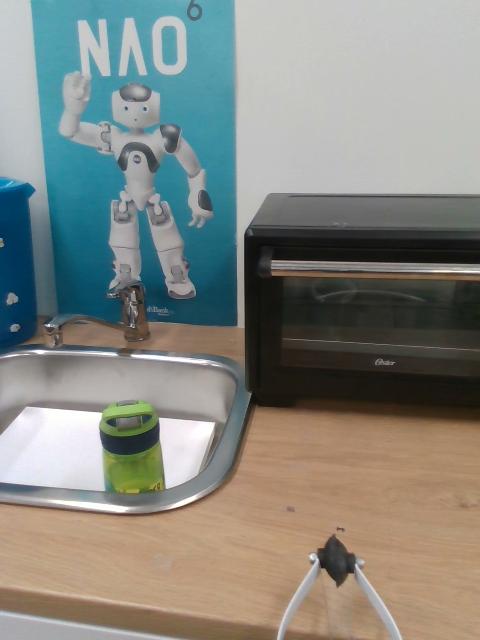} 
    \caption{The full set of observations for one of the demonstrations for the \textit{Bottle to Sink} task.}
    \label{fig:demo_example}
\end{figure}

{
Demonstrations were collected using ROS and a dedicated desktop. A controller was used to issue commands, which were then translated into robot controls (e.g. up and down on the analog stick indicated the arm should move up and down, and left and right translated to moving the arm in and out). The robot continuously provided observation data (RGB-D images, joint state), which was collected at key points specified by the demonstrator via button press. These key points were roughly chosen such that linear interpolation would result in successful completion of of the task. Actions, which we specify as end effector position and orientation in the world frame, were computed using forward kinematics. Demonstrations were validated by replaying them on the robot. As an example, all collected observations for one of the \textit{Bottle to Sink} demonstrations is shown in Figure \ref{fig:demo_example}. 
}

\subsection{Example Grasps}

\begin{figure}[H]
    \centering
    \includegraphics[width=0.2\textwidth]{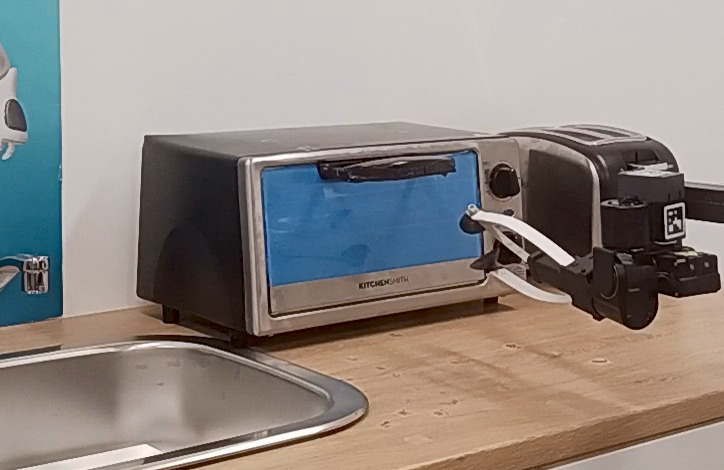} \includegraphics[width=0.2\textwidth]{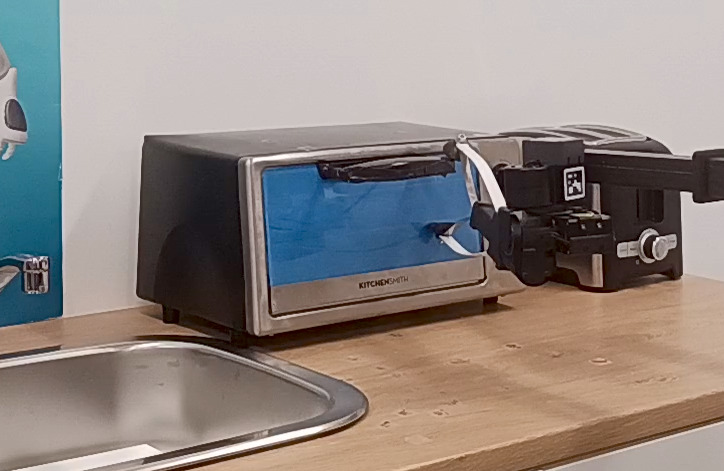}
    \includegraphics[width=0.2\textwidth]{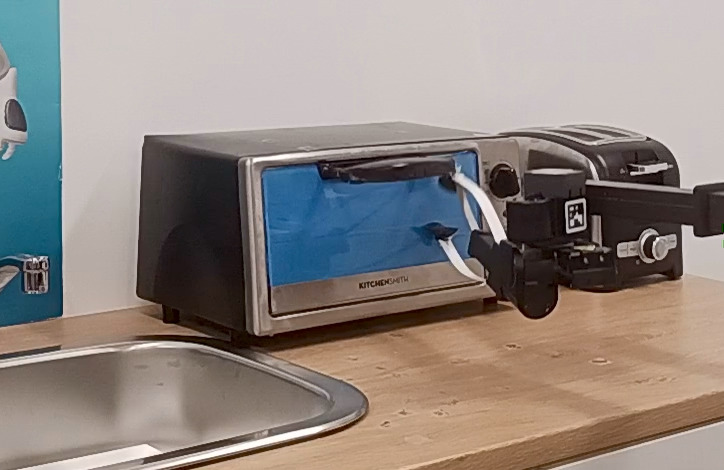}
    \includegraphics[width=0.2\textwidth]{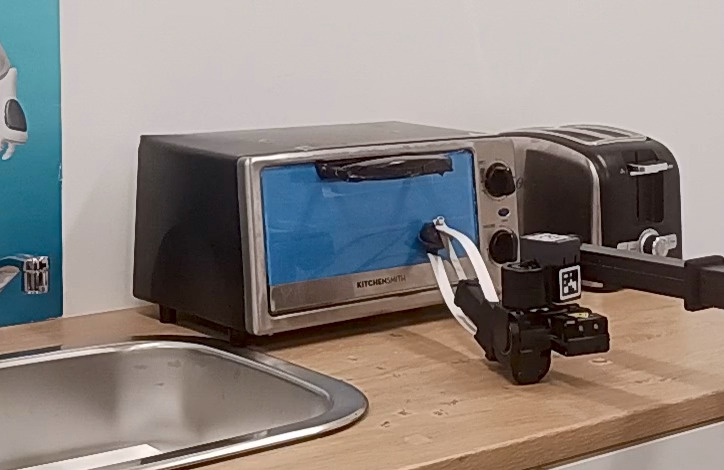} \\
    \includegraphics[width=0.2\textwidth]{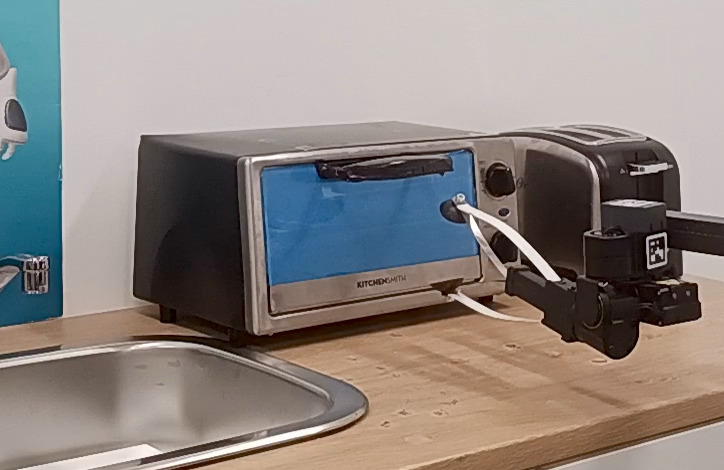}
    \includegraphics[width=0.2\textwidth]{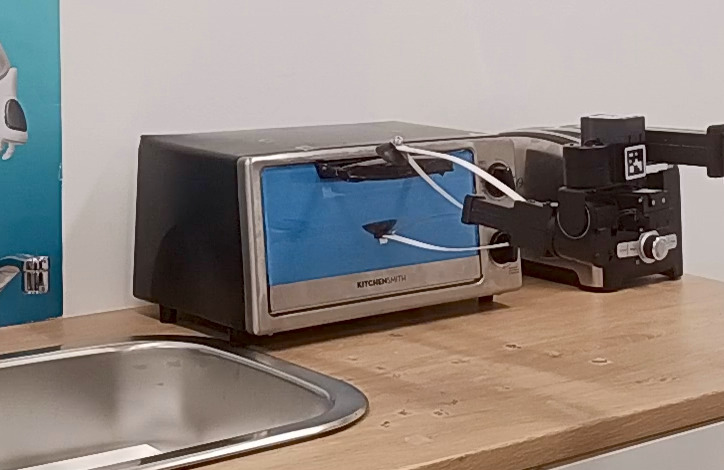}
    \includegraphics[width=0.2\textwidth]{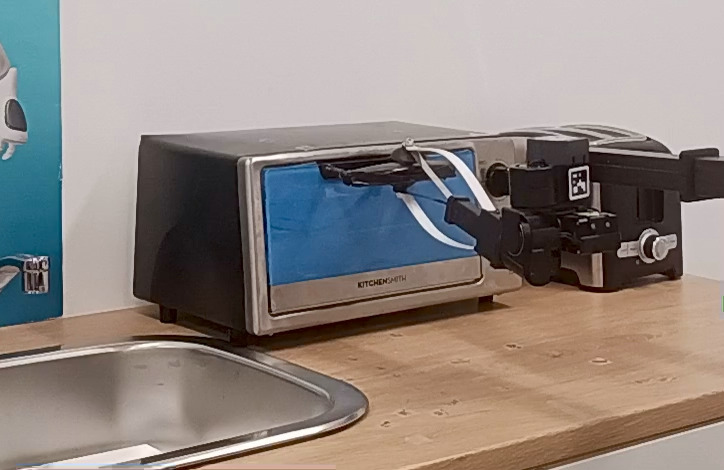}
    \includegraphics[width=0.2\textwidth]{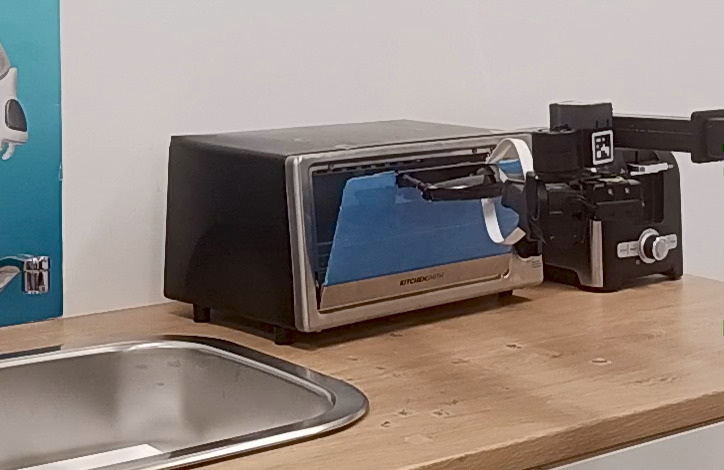}
    \caption{An example of \saner{} executing a re-grasp. In the first row we see the first grasp attempt, in which the robot's gripper slips over the handle. In the second row we see \saner{} pulling back, lining up, and successfully executing the grasp.}
    \label{fig:regrasp}
\end{figure}

\subsection{MRP Attention Visualization}

\subsection{Environment}
\label{sec:environment}
RL methods collect data by taking actions in an environment. To easily adapt to the imitation learning setting, we created an RL-like environment {(according to the OpenAI Gym specification)} that randomly samples transitions from our recorded trajectories. Instead of returning a reward as in a normal RL environment, this environment returns the true action recorded from the demonstration. This is then used by the supervised loss described in Section \ref{sec:abip_loss} in place of the V-trace loss~\citep{espeholt2018_impala} originally used by SANE. We also created a {Gym-like} environment that runs the policy on the robot.

Formulating the imitation learning in this way has a few advantages: first, it makes training on demonstration data use the same interface as running the policy on the robot. It would also allow us to easily extend the method to an active learning setting where rewards from the environment are available. Finally, it allows us to use the baseline implementations available in CORA.

\subsection{\model{} Network Architecture}
\label{appendix:network_details}



\subsubsection{\mpointnet{}}
Each \mpointnet{} model has the following design:
\begin{enumerate}
    \item The first set aggregation module has the following parameters:
    \begin{enumerate}
        \item Furthest point sampling selects 50\% of points to be centers
        \item Each center has 32 neighbors selected within a radius of 0.05m (prioritized by order in the passed in list, not proximity)
        \item All selected neighbors are passed into a 3 layer MLP with a hidden size of 32, an output size of 32, ReLU activations, and using Instance Norm~\citep{ulyanov2016_instancenorm} with no momentum or affine parameters.
        \item The aggregation of encodings is additive.
    \end{enumerate}
    
    \item The second set aggregation module has the following parameters:
    \begin{enumerate}
        \item Furthest point sampling selects 25\% of points to be centers
        \item Each center has 32 neighbors selected within a radius of 0.2m 
        \item All selected neighbors are passed into a 3 layer MLP with hidden sizes of 64, an output size of 128, ReLU activations, and using Instance Norm as above
        \item The aggregation of encodings is additive
        \item After aggregation, the embeddings are passed into another 3-layer MLP with hidden dimension 128 and output dimension 256. 
    \end{enumerate}

    \item The attention network is a 2 layer MLP with hidden dimensions of 32, and an output dimension of 1. 

    \item When computing the locality loss, $\bar{p}$ is detached first. This is to avoid driving $\bar{p}$ towards the unweighted center of the point cloud. 
\end{enumerate}

\subsubsection{Policy Networks}
Each policy head (position, rotation, gripper state, and completion fraction) is a 3 layer MLP with hidden dimension 32 and ReLU activation. 

Inputs into the completion fraction are detached before use; this is because this estimate can be inconsistent between demonstrations, and is not necessary to complete the task. 

{\subsection{Hyperparameters}}
\label{appendix:hyperparameters}

\begin{table}[H]
    \centering
    \begin{tabular}{c|ccccc}
         &  \model{} & No MRP Locality & No Offset Locality & Single Stream & No MRP\\
         \hline
        locality cost: MRP & 100 & 0 & 100 & - & -  \\
        locality cost: action prediction & 10 & 10 & 0 & 10 & 10 \\
        offset dist cost & 100 & 100 & 100 & 100 & 100\\
        dual stream & yes & yes & yes & no & no\\
        uses MRP & yes & yes & yes & yes & no \\
    \end{tabular}
    \caption{Hyperparameters used for ablating \model{}.}
    \label{tab:ablation_hyperparameters}
\end{table}

\begin{table}[H]
    \centering
    \begin{tabular}{cc|ccc}
         & &  CLEAR & SANER & EWC \\
         Common Params & & & & \\
         \hline
        & actor learning rate & 3e-4 & 3e-4 & 3e-4 \\
        & actor cost & 1e5 & 1e5 & 1e5\\
        & batch size & 4 & 4 & 4\\
        & replay ratio & 6 & 6 & -\\
        & augmentation frames & 2500 & 625 * 4 & -\\
        & locality cost: mrp & 200 & 1200 & 100 \\
        & locality cost: offset & 20 & 120 & 10\\
        & offset dist cost & 200 & 1200 & 100\\
         SANER Params & & & & \\
        \hline
        & critic learning rate & - & 1e-4 & -\\
        & activation score cost & - & 1000 & - \\
        & uncertainty cost & - & 100 & -\\
        & activation scale $k_{act}$ & - & 1.0 & -\\
        & lower bound scale & - & 1.0 & -\\
        & upper bound scale & - & 2.0 & -\\
        & source node activation factor ($k_{s,act}$) & - & 0 & -\\
        & source node upper bound factor ($k_{s,ub}$) & - & 100 & -\\
        & new node activation factor ($k_{n,act}$) & - & 100 & -\\
        & new node upper bound factor ($k_{n,ub}$) & - & 5 & -\\
        & activation factor decay ($\gamma_{act}$) & - & 0.99 & -\\
        & upper bound factor decay ($\gamma_{ub}$) & - & 0.995 & -\\
         EWC Params & & & & \\
        \hline
        & Per task frames & - & - & 4000 \\
        & EWC $\lambda$ & - & - & [1000, 10000, 100000] \\
        & Min Frames & - & - & 2000 \\
    \end{tabular}
    \caption{Hyperparameters used for the continual learning experiments.}
    \label{tab:main_hyperparameters}
\end{table}

{Hyperparameters for our experiments are shown in Tables \ref{tab:main_hyperparameters} and \ref{tab:ablation_hyperparameters}. Note that in the single stream ablation, we do not use the MRP stream, instead using the weighted average position from the action prediction stream as the MRP. }

\subsection{Evaluation criteria}
\label{appendix:evaluation}

Scores are reported according to the best behavior observed during the run. E.g. if the robot successfully grasps the bottle and places it in an incorrect location, then attempts a re-grasp and the bottle slips out, the robot would earn a score of 0.6 from the first interaction. 

If the objects ends up in the intended location after a partial grasp (e.g. the bottle is grasped and then pushed into the sink instead of being fully picked up), we still count that as a successful placement. If the object is placed in the correct location (with release), and then perturbed (e.g. the Jello is knocked off of the oven), we still count that as a correct placement.

\textbf{Pick and place tasks}
\textit{Bottle to Sink} and \textit{Bottle from Sink}, 
are scored according to:

\begin{enumerate}
    \item 0.2: Partial grasp (both fingers make contact, but lose contact before raising)
    \item 0.4: Full grasp, partial raise (fingers lose contact during raise)
    \item 0.6: Successful raise, inaccurate placement or early release (e.g. not raised high enough)
    \item 0.8: Successful placement (including opening the gripper), failure to disengage 
    \item 1.0: Full placement and disengagement (hand has released from the object; full retraction not required)
\end{enumerate}

Task specific notes: 
In Bottle To Sink, getting stuck under the oven door counts as "inaccurate placement"

\textbf{Oven interaction tasks}
Open oven and close oven are scored according to:

\begin{enumerate}
    \item 0.2: Partial grasp, no oven door motion
    \item 0.4: Partial oven door movement (<1/4 of full motion)
    \item 0.6: Half pull (~1/2 open)
    \item 0.8: Full door state change, unsuccessful release
    \item 1.0: Full disengagement
\end{enumerate}

If the robot succeeds at opening or closing the oven without the grasp specified in the demonstration, we still count that as success at the task. If the robot releases the handle while its bottom gripper is touching the table, we also count that as a success, since it's hard to open it appreciably more than that, and getting it to stay requires finesse. 

\subsection{\saner{} Activation Score Function}
\label{appendix:sane_score}

As mentioned in Section \ref{sec:saner_critic}, a key goal for our score function is to be sensitive to errors approximately near the tolerance of our task and hardware, but to be less sensitive to the differences between larger errors. 

We chose the $softplus$ function for its two key features: 1) an asymptote at 0, 2) no asymptote to a maximum value. 

To determine the parameters to use with our score function: $softplus(\beta)(a*x+b)$, we set $\beta=10$ and solved such that a distance of 1cm would give a score of 1, and a distance of 5cm would give a score of 0.5 This gives us $a=-12.5$ and $b=1.13$.

{For the orientation metric, we use the quaternion absolute distance and divide by 10, to scale the errors to be comparable to our position errors. This allows us to use the same $softplus$ score function as above. For the gripper, we use the absolute difference in position, and again use the same score function as above.}



\end{document}